\documentclass[letterpaper]{article} 
\usepackage{aaai25}  
\usepackage{times}  
\usepackage{helvet}  
\usepackage{courier}  
\usepackage[hyphens]{url}  
\usepackage{graphicx} 
\usepackage{natbib}  
\usepackage{caption} 
\usepackage{algorithm}
\usepackage{algorithmic}
\usepackage{amsmath}
\usepackage{xcolor}
\usepackage{cite}
\usepackage{booktabs}
\usepackage{bbding}
\usepackage{multirow}
\usepackage{verbatim}
\usepackage{url}
\usepackage{array}
\usepackage{amsmath,amsfonts}

\usepackage{newfloat}
\usepackage{listings}
\DeclareCaptionStyle{ruled}{labelfont=normalfont,labelsep=colon,strut=off} 
\floatstyle{ruled}
\newfloat{listing}{tb}{lst}{}
\floatname{listing}{Listing}
\title{AFiRe: Anatomy-Driven Self-Supervised Learning for Fine-Grained Representation in Radiographic Images}

\author{
    Yihang Liu \textsuperscript{\rm 1},
    Lianghua He \textsuperscript{\rm 1}\textsuperscript{\rm 2}\thanks{Corresponding author.}, 
    Ying Wen \textsuperscript{\rm 3}, 
    Longzhen Yang \textsuperscript{\rm 1}, 
    Hongzhou Chen \textsuperscript{\rm 1}
}
\affiliations{
    \textsuperscript{\rm 1} School of Computer Science and Technology, Tongji University, Shanghai, China.\\
    \textsuperscript{\rm 2} Shanghai Eye Diseases Prevention and Treatment Center, Shanghai Eye Hospital, Shanghai, China.\\
    \textsuperscript{\rm 3} School of Communication and Electronic Engineering, East China Normal University, Shanghai, China.\\

    \{2111131, helianghua, yanglongzhen, chenhonghzou\}@tongji.edu.cn, ywen@cs.ecnu.edu.cn
}

\usepackage{bibentry}

\begin{document}

\maketitle

\begin{abstract}
Current self-supervised methods, such as contrastive learning, predominantly focus on global discrimination, neglecting the critical fine-grained anatomical details required for accurate radiographic analysis. To address this challenge, we propose an \textbf{A}natomy-driven self-supervised framework for enhancing \textbf{Fi}ne-grained \textbf{Re}presentation in radiographic image analysis (AFiRe). The core idea of AFiRe is to align the anatomical consistency with the unique token-processing characteristics of Vision Transformer. Specifically, AFiRe synergistically performs two self-supervised schemes: (i) Token-wise anatomy-guided contrastive learning, which aligns image tokens based on structural and categorical consistency, thereby enhancing fine-grained spatial-anatomical discrimination; (ii) Pixel-level anomaly-removal restoration, which particularly focuses on local anomalies, thereby refining the learned discrimination with detailed geometrical information.
Additionally, we propose Synthetic Lesion Mask to enhance anatomical diversity while preserving intra-consistency, which is typically corrupted by traditional data augmentations, such as Cropping and Affine transformations.
Experimental results show that AFiRe: (i) provides robust anatomical discrimination, achieving more cohesive feature clusters compared to state-of-the-art contrastive learning methods; (ii) demonstrates superior generalization, surpassing 7 radiography-specific self-supervised methods in multi-label classification tasks with limited labeling; and (iii) integrates fine-grained information, enabling precise anomaly detection using only image-level annotations. 
\begin{links}
 \link{Code}{https://github.com/LYH-hh/AFiRe}
\end{links}

\end{abstract}

%

\section{Introduction}
Contrastive learning (CL) has emerged as a powerful method in self-supervised learning (SSL), demonstrating its effectiveness without relying on expert annotations \cite{swav, dino}. 
In natural image analysis, prevailing CL methods like SimCLR \cite{SimCLR} and MoCo \cite{moco} primarily focus on global discrimination. 
Explicitly, they consider the entire image and its transformations as positive pairs, while treating different images as negative pairs \cite{Contrastive-instance-discrimination, Contrastive-pretext-invariant}.
Although these models successfully capture high-level representations across diverse images and achieve invariance to salience changes, their efficacy is limited when applied to radiographic images \cite{PCRL, Dira}. 

Natural images exhibit invariance of foreground objects across diverse backgrounds, which directs contrastive tasks to emphasize image-level discrimination \cite{Contrastive-pretext-invariant}. In contrast, radiographic images present clinical salience dispersed throughout the image \cite{mlip}, necessitating the model's attention to fine-grained discrimination, including the distribution of densities, the arrangement of tissues, and the presence of specific pathologies \cite{AnaXNet, 3D_anatomy}. For example, faint opacities may indicate early signs of pneumonia, while fine reticular patterns can suggest interstitial lung disease. Therefore, incorporating fine-grained details, such as detailed anatomical structures and complex spatial relationships \cite{Dirav2, discrimination_pami}, into high-level representations is essential for accurately identifying normal anatomy and detecting various lesions in radiographic images. 

\begin{figure}
  \centering
  \includegraphics[width=1\linewidth]{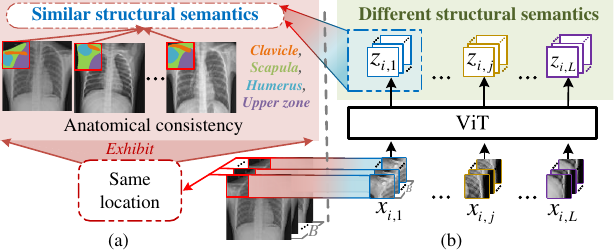}
  \caption{\textbf{Conception of the proposed method.} (a) Local radiographic structures at the same location exhibit anatomical consistency.
  (b) By aligning anatomical consistency with the token-processing characteristics of ViT, tokens at the same position within a batch share similar structural semantics, while those at different positions convey distinct ones.}
  \label{fig:first}
\end{figure}

Recent advances in medical contrastive learning (MCL) have incorporated fine-grained information by focusing on region discrimination \cite{discrimination_pami, MLVICX, mlip}. These approaches aim to embed semantically consistent features while distinguishing them from features with distinct semantics \cite{Adamv2}. However, the variability of diseases makes it difficult to comprehensively capture pathological semantics, and additional resources are required to determine the relevant areas (such as pre-selection \cite{pre_roi}). Another effective method for enriching fine-grained information is to integrate the pixel restoration with MCL \cite{PCRL,Dirav2}. However, these methods typically exhibit a deficiency in anatomical semantics, as they restore pixel-level content directly from latent representations. This plain pixel information limits the model’s capability to focus on salient clinical patterns.

In this paper, to circumvent the challenge posed by heterogeneous anomalies, which potentially leads to insufficient learning of semantic information, we highlight that \textit{comprehensive learning of normal anatomical patterns is comparatively easier.} This advantage stems from the significant semantic similarity observed in normal radiographic images \cite{SQUID}, which are characterized by anatomical consistency and stable focal regions across different individuals (Fig. \ref{fig:first}(a)). Such similarity facilitates the effective decoupling of underlying anomalies from normal structures, thereby enhancing pathological diagnosis performance. 
Additionally, to avoid region pre-selection, we hypothesize that \textit{each distinct token in Vision Transformers (ViTs) \cite{vit} explicitly represents the predominant information of specific local anatomical structures}.
This hypothesis is inspired by SelfPatch \cite{Patch-level}, 
which asserts that ViTs have inherent architectural advantages for enhancing visual representations through the processing of discrete image tokens. 
Summarily, our core idea is to \textit{\textbf{align anatomical consistency with the unique token-processing characteristics of ViT}} to enhance fine-grained radiographic representation (Fig. \ref{fig:first}(b)).
These alignments are also pivotal in anomaly-specific restoration for preserving pixel-level anatomy-associated (\textit{i.e.}, geometrical) information.

Based on the analysis above, we propose a novel \textbf{A}natomy-driven self-supervised framework to enhance \textbf{Fi}ne-grained \textbf{Re}presentation in radiographic images (\textbf{AFiRe}). Equipped with the designed Synthetic Lesion Masks (SLMs), an anatomy augmentation inspired by AnatPaste \cite{Anatpaste}, AFiRe aligns ViT tokens with local anatomical structures via synergistically performing two SSL tasks: Token-wise anatomy-guided contrastive learning and Pixel-level anomaly-removal restoration.

For the contrastive learning task, instead of analyzing entire images, AFiRe processes each disjoint token in ViT as an independent sample. By maximizing mutual information among these tokens within a batch, this task learns fine-grained structural invariance and discriminative semantics based on structural and categorical consistency. To guide the model in comprehensively learning the normal anatomical patterns, we introduce a group of spatial-aware prototypes in this task. These prototypes represent the distribution of different anatomical structures, serving as pseudo-cluster assignments for the predicted token probabilities.
For the pixel restoration task, we strategically focus on restoring specific abnormal tokens augmented by SLMs. Concretely, we remove these abnormal tokens by substituting them with trainable masked tokens to restore their corresponding normal pixels. Hence, the latent representations retain both normal and abnormal information while preserving geometrical details closely associated with various anatomical structures.

Our extensive experiments demonstrate that AFiRe enhances the model's capacity to capture fine-grained discriminative information from normal radiographic images, facilitating a more robust representation across different anatomic structures. 
Compared to different supervised and self-supervised benchmarks, AFiRe achieves superior performance in multi-label classification tasks, indicating its generalization in analyzing real disease images. Furthermore, AFiRe has observable advantages in three anomaly detection tasks, outperforming the state-of-the-art counterparts.
Overall, our contributions are as follows:

\begin{itemize}
	\item We design a Token-wise anatomy-guided contrastive learning SSL task, equipped with spatial-aware prototypes to integrate fine-grained anatomical discrimination based on structural and categorical consistency.
	\item We design a Pixel-level anomaly-removal restoration SSL task to preserve detailed geometrical information closely associated with various anatomical structures.
    \item We introduce Synthetic Lesion Mask, an efficient anatomical data augmentation technique, to enhance anatomical diversity while preserving intra-consistency.
    \item We propose an Anatomy-driven self-supervised framework that synergistically optimizes the aforementioned tasks to achieve robust radiographic representation with fine-grained anatomical information.
\end{itemize}

\begin{figure*}
  \centering
  \includegraphics[width=1\linewidth]{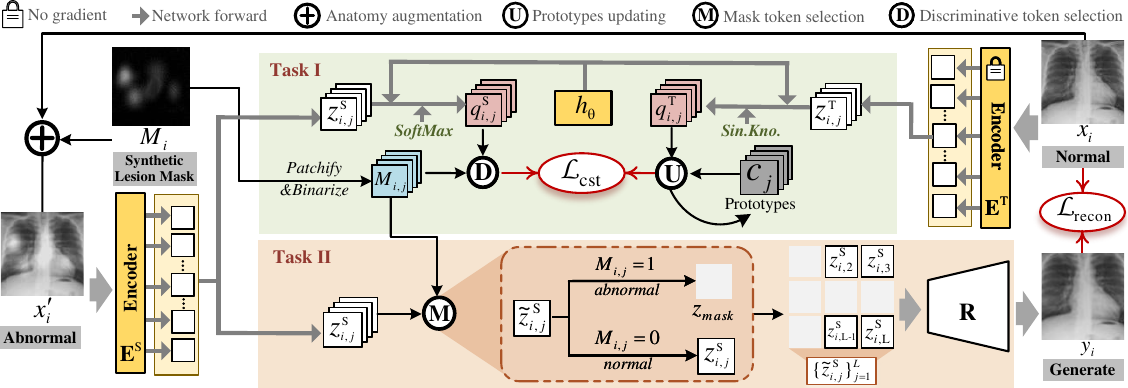}
  \caption{\textbf{Overview of the proposed AFiRe}. It synergistically performs two self-supervised proxy tasks: Token-wise anatomy-guided contrastive learning (\textbf{Task \uppercase\expandafter{\romannumeral1}}) and Pixel-level anomaly-removal restoration (\textbf{Task \uppercase\expandafter{\romannumeral2}}). For each normal input $x_i$, we perturb it using the designed Synthetic Lesion Mask ($M_i$) to produce abnormal input $x^\prime_i$. 
  In Task \uppercase\expandafter{\romannumeral1}, a group of spatial-aware prototypes, updated by the teacher network's output, serve as pseudo-cluster labels to maximize alignment among tokens from student networks belonging to the same class or structure. In Task \uppercase\expandafter{\romannumeral2}, the restoration target particularly focuses on the abnormal tokens from augmented pairs of normal radiographic images by substituting them with mask tokens in the latent space.
  }
  \label{fig:main}
\end{figure*}
\section{RELATED WORK}
\label{sec:formatting}
SSL has become a prominent topic in medical image analysis, enabling the extraction of meaningful representations directly from data without the need for explicit disease-specific labels. In this section, we review recent advancements that are most relevant to our proposed AFiRe method.

\textbf{Medical Contrastive Learning.} 
This approach typically uses encoders to cluster instances within the pseudo-classes \cite{MoCo-CXR, MICLe_contrast_CXR, Medaug_contrast_CXR}. Recent research, including C2L \cite{C2L} adapts general contrastive methods to learn distinctive patterns across various medical images. Adamv1 \cite{Adamv1} applies hierarchical contrastive learning, capturing anatomy in a coarse-to-fine manner. Adamv2 \cite{Adamv2} expands it by enhancing radiographic representation with localizability, composability, and decomposability. While these methods show effective applications in medical tasks, they remain focused on image-level discrimination based on the input, neglecting the fine-grained anatomical information that could be aligned with token-level representations.

\textbf{Medical Restorative Learning.} 
This approach reconstructs original images from corrupted versions, enhancing pixel-level details to identify normal anatomy and various abnormalities \cite{Dirav2}. Key advancements include Model Genesis (MG) \cite{MG}, which uses image restoration tasks on unlabeled medical images; TransVW \cite{TransVW}, which defines a “visual word” as a recurring anatomical segment across images. SQUID \cite{SQUID} leverages space-aware memory queues to capture spatial correlations and consistent anatomical structures in chest images, thereby demonstrating effectiveness in anomaly detection through unsupervised learning. AnatPaste \cite{Anatpaste} introduces anatomy-aware pasting augmentation, generating synthetic images to distinguish real normal images with a one-class classifier under SSL. These methods have shown promise in radiographic representation learning by preserving detailed patterns; however, they are often suboptimal for radiography classification due to limited high-level discrimination.

\textbf{Medical Synergistic Learning.} 
To harness complementary advantages, synergistic SSL approaches have been proposed for radiographic representation learning. DiRA \cite{Dira} combines discriminative, restorative, and adversarial learning to capture complementary visual information, enhancing fine-grained semantic representation learning. PCRL \cite{PCRL} incorporates preservation mechanisms to reconstruct diverse image contexts, refining the representations derived from contrastive methods.

While these models offer progress, the proposed AFiRe advances the comprehensive integration of anatomical information by introducing Token-Wise Anatomy-Guided Contrastive Learning and Pixel-Level Anomaly-Removal Restoration, achieving both fine-grained high-level discrimination and preservation of pixel-level anatomical geometry.

\section{METHOD}
To enhance fine-grained anatomical representation in radiographic image analysis, we propose an anatomy-driven self-supervised framework as depicted in Fig. \ref{fig:main}. 
Our framework employs a siamese ViT architecture to extract features from real normal and generated abnormal images by introducing the Synthetic Lesion Mask. Based on the alignment of anatomical consistency with the unique token-processing
characteristics of ViT, this framework synergistically performs two SSL tasks: (i) \textbf{Task \uppercase\expandafter{\romannumeral1}}, the Token-wise anatomy-guided contrastive learning, aims to enhance fine-grained discriminative information by maximizing mutual information among individual image tokens within the same class or anatomical structure; and (ii) \textbf{Task \uppercase\expandafter{\romannumeral2}}, the Pixel-level anomaly-removal restoration, aims to incorporate geometric-based anatomical details by optimizing the alignment of abnormal tokens between the normal image and its reconstructed counterpart at the pixel level.

In the following, we introduce each component individually and then discuss the synergistic training scheme.

\subsection{Feature Extraction with Siamese Network}
\label{sec:encoder}
As shown in Fig. \ref{fig:main}, the overall structure of the AFiRe comprises a student branch and a teacher branch.
In our training scheme, the student network $\mathbf{E}^\text{S}$, parameterized by $\theta^\text{S}$, is trained to match the spatial-aware prototypes updated by the output of the teacher network $\mathbf{E}^\text{T}$, parameterized by $\theta^\text{T}$. Both networks share identical ViT architectures. During the pre-training phase, only $\theta^\text{S}$ is updated via back-propagation, while $\theta^\text{T}$ is updated using an exponential moving average (EMA) \cite{moco, dino} of $\theta^\text{S}$ as follows: 
\begin{align}
\theta^\text{T} \leftarrow \lambda\theta^\text{T} + (1-\lambda)\theta^\text{S},
\end{align}
where $\lambda$ follows a cosine schedule from 0.99 to 1. 

To leverage the anatomical consistency observed in normal radiographic images \cite{SQUID}, we incorporate two types of input data during the pre-training stage: (i) Real normal images $\{x_i\}_{i=1}^B$, which provide a baseline to establish the distribution of typical anatomical structures, and (ii) Synthetic abnormal images $\{x_i^{\prime}\}_{i=1}^B$, which integrate category discriminative information by simulating pathologies using Synthetic Lesion Masks.
Here, $B$ represents the batch size, and the images $x_i$ and $x^\prime_i$ are encoded by the teacher network $\mathbf{E}^\text{T}$ and the student network $\mathbf{E}^\text{S}$, respectively. 

\textbf{\textit{Alignment anatomical consistency with ViT tokens.}} 
ViTs process input images as sequences of non-overlapping patches, with the extracted features $\{z_{i,j}^\text{T}\}_{j=1}^L$ and $\{z_{i,j}^\text{S}\}_{j=1}^L$ (white blocks in Fig \ref{fig:main}) corresponding to token-wise representations, where $L$ is the length of image token sequences. 
Each $z_{i,j}$ represents the $j$-th positional token, corresponding to the $j$-th local anatomical structure in the $i$-th image. In the subsequent Task \uppercase\expandafter{\romannumeral1} and Task \uppercase\expandafter{\romannumeral2}, we explicitly align the anatomical consistency exhibited in $\{z_{i,j}^\text{T}\}_{i=1}^B$ and $\{z_{i,j}^\text{S}\}_{i=1}^B$ with the unique token-processing characteristics of ViTs to enhance fine-grained radiographic representation.

\textbf{\textit{Anatomy Augmentation via Synthetic Lesion Mask.}} Traditional image data augmentations, such as geometric transformations including Rotation, Cropping, Affine, and Flipping, tend to corrupt anatomical consistency due to altered spatial relationships. Popular pasting-based augmentations like CutPaste \cite{cutpaste} are designed to introduce variations in images without causing geometric distortions. However, these techniques often produce abnormalities with clear boundaries, which fail to capture the effective characteristics of real anomalies. Recently, the anatomy-aware pasting (AnatPaste) augmentation technique \cite{Anatpaste} has been developed to create anomalies with anatomical fidelity by introducing abnormal shadows within the extracted lung regions. In this section, we improve AnatPaste in an efficient way to generate more random anomalies by introducing the Synthetic Lesion Mask (SLM).

Unlike AnatPaste, which uses fixed anomalous sizes and textures, SLM implements a dynamic anatomical augmentation to efficiently synthesize various pathological anomalies within a single normal image. 
Specifically, SLM randomly selects a highlighted region $\eta$, binarized by \cite{otsu}, from $x_i$ to serve as universal texture noise. For the size of $x_i \in \mathbb{R}^{224 \times 224}$, we simulate lesion sizes ranging from small nodules to large consolidations by randomly resizing $\eta$ to $\mathbb{R}^{W \times H}$, where $W, H \in [16,64]$. We then define the lesion shapes as irregular ovals: 
\begin{align}
  &\delta(w,h)=\text{exp}(-\frac{\rho(w,h)}{\gamma}^2), \\
  &\rho(w,h)=\sqrt{(w-\frac{W}{2})^2+(h-\frac{H}{2})^2}.
\end{align}
In the shape $\delta(w,h)\in \mathbb{R}^{W\times H}$, the value at each pixel $(w,h)$ gradually decreases from the center to the edge with the radius $\gamma$. We integrate $\eta$ with $\delta(w,h)$ and randomly put them onto a zero-valued map of the same size as $x_i$. The process of creating a group of SLMs $\{M^n_i\}_{n=1}^N$ for each $x_i$ is formulated as follows: 
\begin{align}
\{M^n_i\}_{n=1}^N = \{\sum_{r=1}^R{\eta^r \cdot \delta^n(w,h)}\}_{n=1}^N, 
\end{align}
where $N$ is the number of SLMs, and $R$ represents the number of abnormal regions in a single $M_i^n$. Empirically, we set $R\in [1,4]$. The abnormal images can be obtained by: 
\begin{align}
\{x_i^{\prime,n}\}_{n=1}^N = \{x_i\oplus M^n_i\}_{n=1}^N,
\end{align}
where $\oplus$ is the element-wise additional operation to augment the anatomy.

\subsection{Token-wise Anatomy-guided Contrastive Learning}
\begin{figure}
  \centering
  \includegraphics[width=1\linewidth]{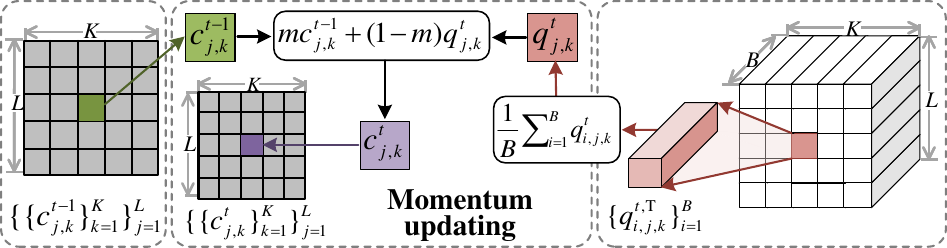}
  \caption{\textbf{Updating process of the spatial-aware prototypes}. The cluster assignment of $\mathbf{E}^\text{T}$ is used for updating the spatial-aware prototypes.}
  \label{fig:prototypes}
\end{figure}

\begin{figure}
  \centering
  \includegraphics[width=1\linewidth]{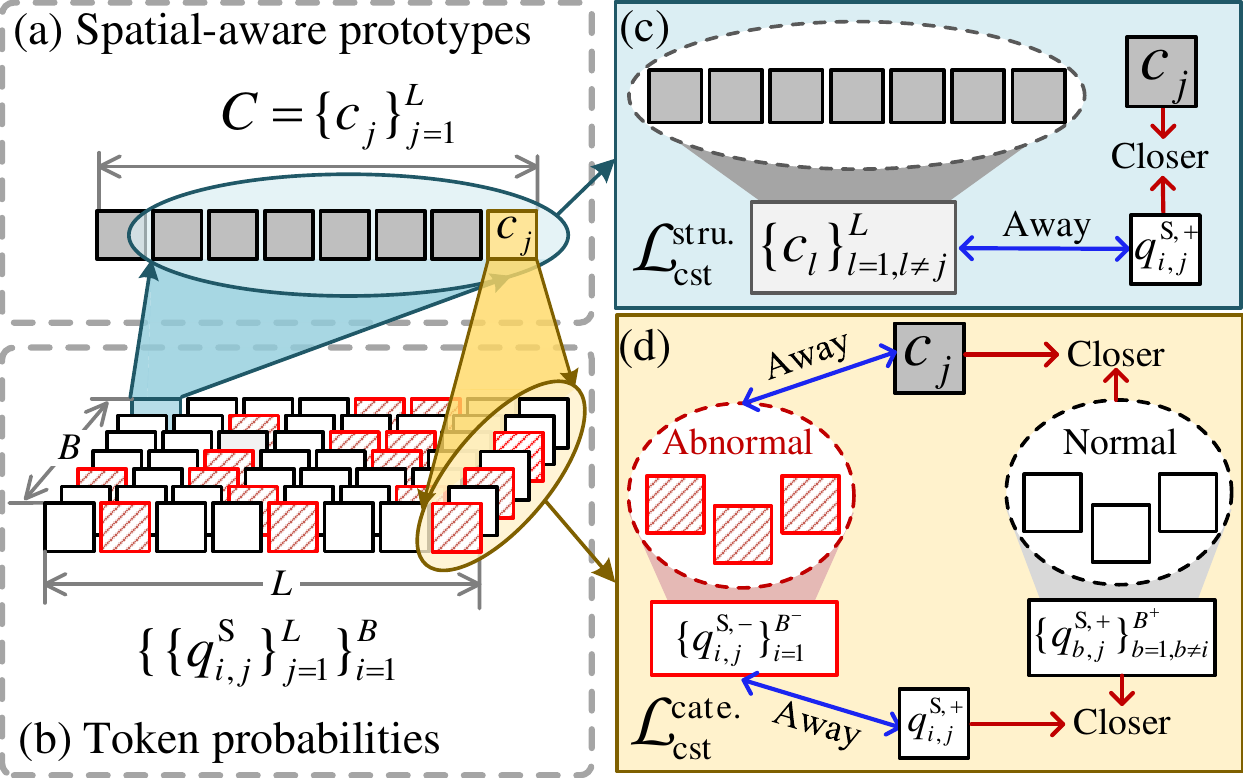}
  \caption{\textbf{Token-wise anatomy-guided contrastive learning}. $\mathcal{L}_{\text{cst}}^\text{stru.}$ and $\mathcal{L}_{\text{cst}}^\text{cate.}$ correspond to the structure-consistency and category-consistency contrastive losses, respectively.}
  \label{fig:cst_loss}
\end{figure}

Instead of contrasting the image-level representation, in this task, we consider each token as an individual sample.
Specifically, we propose Token-wise anatomy-guided contrastive learning (\textbf{Task \uppercase\expandafter{\romannumeral1}}), which maximizes the mutual information between the probability distribution of normal image tokens and their respective structural prototypes (Fig. \ref{fig:prototypes}) by integrating structure-consistency and category-consistency contrastive losses (Fig. \ref{fig:cst_loss}).

\textbf{\textit{Token Probabilities Prediction.}}
To mitigate the impact of individual anatomical differences, our contrastive learning task focuses on cluster assignments rather than direct feature comparisons.
Specifically, we employ a projection head ($h_\theta$) to map each token from the $dim$-dimensional feature ($\{z_{i,j}^\text{S}\}$ and $\{z_{i,j}^\text{T}\}$) to $K$-dimensional vectors as shown in Task \uppercase\expandafter{\romannumeral1} of Fig. \ref{fig:main}.
The student token probability distributions $q_{i,j}^\text{S} \in \mathbb R^{K}$ are obtained by normalizing these $K$-dimensional vectors with a $SoftMax$ function, while the teacher token probability distributions $q_{i,j}^\text{T} \in \mathbb R^{K}$ are computed via the Sinkhorn-Knopp ($Sin.Kno.$) algorithm \cite{Sinkhorn} to avoid trivial solutions where all samples collapse into a single representation.
Here, the $q_{i,j,k}$, where $k\in[1, K]$, denotes the predicted probability that the $j$-th token of the $i$-th image belongs to class $k$ within its structural cluster assignment. 

\textbf{\textit{Discriminative Token Selection.} }
To achieve consistency among normal anatomic structures, it is important to identify probability distributions in normal and abnormal categories.
Synthetic Lesion Mask $M_i$ is regarded as the token label to distinguish normal and abnormal image tokens within the set $\{q_{i,j}^\text{S}\}_{j=1}^L$, corresponding to the `\textbf{D}' operation in Fig \ref{fig:main}. 
Specifically, $M_i$ is patchified with the same patch size as $x_i$, resulting in a sequence $\{M_{i,j}\}_{j=1}^L$. We define the threshold as the average pixel value of $M_i$. Therefore, for each patch, $M_{i,j}=1$ indicates that the patch's average pixel value exceeds the threshold; otherwise, $M_{i,j}=0$. Thus, the image tokens can be collected in two categories:
\begin{equation}
\begin{aligned}
  \{q_{i,j}^{\text{S},+}\} = \{q_{i,j}^\text{S}|M_{i,j}=0\}, \;
  \{q_{i,j}^{\text{S},-}\} = \{q_{i,j}^\text{S}|M_{i,j}=1\},
\end{aligned}
\end{equation}
where $q_{i,j}^{\text{S},+}$ and $q_{i,j}^{\text{S},-}$ indicate the normal and abnormal token probability, respectively.

\textbf{\textit{Spatial-aware Prototypes Construction and Update.}}
To learn comprehensive semantics of various normal anatomical structures, we adopt the concept of clustering prototypes from SwAV \cite{swav}. Our approach diverges significantly from SwAV by introducing a group of distinct prototype vectors (the gray blocks in Fig. \ref{fig:main}), allowing the model to more effectively represent the probabilistic distribution across different anatomies. In our pre-training stage, these prototypes act as the pseudo cluster assignments for $q_{i,j}^\text{S}$, thereby guiding the update of gradient for $\theta^\text{S}$. 

In practice, these vectors are learned through momentum updating rather than back-propagation (the `\textbf{U}' operation in the Fig \ref{fig:main}). We define a 2D prototype matrix $\mathbf{C} = \{\{c_{j,k}\}_{k=1}^K\}_{j=1}^L \in \mathbb R^{L\times K}$ as shown in Fig. \ref{fig:prototypes}. This matrix indicates that each distinct token in the image is described by a $K$-dimensional vector. 
At time $t$, the average probability score for the specific $j$-th token $\{q_{i,j,k}^{t, \text{T}}\}_{i=1}^B$ in the $k$-th class is calculated as $q_{j,k}^t=\frac{1}{B}\sum_{i=1}^B (q_{i,j,k}^{t, \text{T}})$ (the pink block in Fig. \ref{fig:prototypes}). The updating of previous prototypes $c_{i,k}^{t-1}$ (the green block in Fig. \ref{fig:prototypes}) can be formulated as: 
\begin{align}
\begin{cases}
c^t_{j,k}=q_{j,k}^t, & \text{if } t=1 \\
c^t_{j,k}=mc^{t-1}_{j,k}+(1-m)q_{j,k}^t, & \text{if } t>1.
\end{cases}
\end{align}
This process ensures that the prototype matrix $\mathbf{C}$ is gradually updated with the predicted probabilities over time, balancing between the historical values and the new incoming data through the momentum parameter $m$.

\textbf{\textit{Structure-consistency Contrastive Loss Formulation.}}
To encode more fine-grained structural discriminative information, this contrastive loss is designed to maintain invariance within the same anatomical structure and increase differentiation among various anatomical structures. Specifically, given the normal token probabilities $Q_{i}^{\text{S},+}=\{q_{i,j}^{\text{S},+}\}_{j=1}^L$ in the $i$-th image (such as the blue block in Fig. \ref{fig:cst_loss} (b)) and the spatial-aware prototypes $C=\{c_{j}\}_{j=1}^L$, we maximize the mutual information as follow:
\begin{align}
  &I(Q_{i}^{\text{S},+}; C) =\sum_{q_{i,j}^{\text{S},+}} \sum_{c_j} P(c_{j}, q_{i,j}^{\text{S},+}) \log \frac{P(c_j|q_{i,j}^{\text{S},+})}{P(c_j)}.
  \label{eq:MI}
\end{align}
Directly optimizing $I(Q_{i}^{\text{S},+}; C)$ is a challenging task. Alternatively, we define the joint probability by considering the likelihood that a token $q_{i,j}^{\text{S},+}$ belongs to a cluster represented by $c_{j}$. The conditional probability is approximated by the normalized similarity between them. This process can be formulated as follows: 
\begin{align}
&P(c_{j}, q_{i,j}^{\text{S},+})=P(c_{j}| q_{i,j}^{\text{S},+})P(q_{i,j}^{\text{S},+}), \label{eq:joint}\\
&P(c_{j}| q_{i,j}^{\text{S},+})=\frac{\text{sim}(q_{i,j}^{\text{S},+}, c_j)}{\sum_{l=1}^L\text{sim}(q_{i,j}^{\text{S},+}, c_l)},
\label{eq:conditional}
\end{align}
where $\text{sim}(.,.)$ is the similarity function. 
Next, we plug Eq. \ref{eq:joint} and Eq. \ref{eq:conditional} into Eq. \ref{eq:MI} and simplify it to obtain: 
\begin{align}
I(Q_{i}^{\text{S},+}; C) = \mathbb{E}_{P(Q_{i}^{\text{S},+}, C)} \left[ \log \frac{\text{sim}(q_{i,j}^{\text{S},+}, c_j)}{\sum_{l=1}^L \text{sim}(q_{i,j}^{\text{S},+}, c_l)} \right].
\end{align}
Appendix A offers additional details for simplification. 

Therefore, $I(q_{i,j}^{\text{S},+}; c_{j}) \propto \text{sim}(q_{i,j}^{\text{S},+}, c_j)$. By considering all the $i \in [1,B^+]$, and $j \in [1, L]$, we maximize $I(Q_{i}^{\text{S},+}; C)$ by proposing the structure-consistency contrastive loss: 
\begin{align}
\mathcal{L}_{\text{cst}}^\text{stru.} = -\frac{1}{LB^+}\sum_{j=1}^L \sum_{i=1}^{B^+} \log \frac{f(q_{i,j}^{\text{S},+}, c_j)}{\sum_{l=1}^Lf(q_{i,j}^{\text{S},+}, c_l)}, 
\label{eq:strue_loss}
\end{align}
where $f(q,c)=\exp(\text{H}(q,c)/\tau)$ and  $B^+$ is the number of $q_{i,j}^{\text{S},+}$. Here, $\tau$ is the temperature parameter and $\text{H}(q,c) = -\sum\nolimits_{k=1}^K q(k)\log c(k)$, which is the cross-entropy function to measure the similarity between two probabilities. The positive and negative pairs in $\mathcal{L}_{\text{cst}}^\text{stru.}$ are defined as depicted in Fig. \ref{fig:cst_loss} (c). We pull a specific normal image token $q_{i,j}^{\text{S},+}$ and its corresponding structural prototype $c_j$ closer while pushing other prototypes $\{c_l\}_{l=1}^L$, where $l \neq j$, away. 

\textbf{\textit{Category-consistency Contrastive Loss Formulation.}}
To effectively decouple pathological information from the normal distribution, we further investigate the consistency among normal image tokens within a batch size $B$ at the same position $j$. Particularly, given the specific $c_j \in C$  (the yellow block in
Fig. \ref{fig:cst_loss} (a)) and $q_{i,j}^{\text{S},+}, q_{b,j}^{\text{S},+}\in Q_{j}^{\text{S},+}$, we maximize the joint mutual information as follows:
\begin{equation}
\begin{aligned}
&I(C, Q_{j^b}^{\text{S},+}; Q_{j^i}^{\text{S},+}) = I(C; Q_{j^i}^{\text{S},+}) + I(Q_{j^i}^{\text{S},+};Q_{j^b}^{\text{S},+}|C),
\label{eq:MI_cate}
\end{aligned}
\end{equation}
where $i, b \in [1, B^+]$ and $b \neq i$.
Since $I(C; Q_{j^i}^{\text{S},+})$ has been optimized through $\mathcal{L}_{\text{cst}}^\text{stru.}$, maximizing Eq. \ref{eq:MI_cate} only needs to optimize the second term, which represents the unique mutual information between $q_{b,j}^{\text{S},+}$ and $q_{i,j}^{\text{S},+}$. Thereby, we introduce a negative term, \textit{i.e.}, reducing the similarity between ${q_{i,j}^{\text{S},-}}$ and $c_{j}$, to constrain $I(Q_{j^i}^{\text{S},+};Q_{j^b}^{\text{S},+}|C)$. Mathematically, similar to Eq. \ref{eq:strue_loss}, we propose the category-consistency contrastive loss $\mathcal{L}_\text{cst}^\text{cate.}$ and introduce positive and negative terms as depicted in Fig. \ref{fig:cst_loss} (d): 
\begin{align}
&\Delta({pos}) = \sum\nolimits_{b=1}^{B^+}f(q_{i,j}^{\text{S},+}, q_{b,j}^{\text{S},+}), \label{eq:pos_cate}\\
&\Delta(neg) = \sum\nolimits_{b=1}^{B^-}f(q_{i,j}^{\text{S},+}, q_{b,j}^{\text{S},-}) + f(q_{i,j}^{\text{S},-}, c_j),
\label{eq:neg_cate}
\end{align}
where $B^-$ is the number of $q_{i,j}^{\text{S}}$ and $B^+ + B^- =B$.
Therefore, the $\mathcal{L}_\text{cst}^\text{cate.}$ can be formulated as follows:
\begin{align}
  \mathcal{L}_{\text{cst}}^\text{cate.} = -\frac{1}{LB}\sum_{j=1}^L \sum_{i=1}^B 
  \log \frac{\Delta(pos)}{\Delta(pos) + \Delta(neg)}.
\end{align}
Appendix B provides detailed proofs.

Notably, given the inherent diversity and irregularity of pathologies in radiographic images, enforcing consistency among abnormal tokens could limit the model's ability to generalize and recognize pathological semantic information (ablation experiments are posted in Appendix F). 
Therefore, in the proposed $\mathcal{L}_{\text{cst}}^\text{cate.}$, we disregard intra-class similarity for the $I(Q_{jb}^{\text{S},-}, Q_{ji}^{\text{S},-})$, where $q_{i,j}^{\text{S},-}, q_{b,j}^{\text{S},-}\in Q_{j}^{\text{S},-}$, to accommodate the heterogeneity of anomalies.

Summarily, in the proposed Token-wise anatomy-guided contrastive task, the loss function is formulated as follows:
\begin{align}
  \mathcal{L}_{\text{cst}} =\mathcal{L}_{\text{cst}}^\text{stru.}+\mathcal{L}_{\text{cst}}^\text{cate.}.
\end{align}

\subsection{Pixel-level Anomaly-removal restoration}
To enhance the learned discrimination with detailed geometrical information, we propose the pixel-level anomaly-removal restoration, as illustrated in the \textbf{Task \uppercase\expandafter{\romannumeral2}} of Fig. \ref{fig:main}.
Unlike conventional methods that apply image restoration across the entire latent feature space, this task concentrates on localized anomalies by restoring image features that exclude abnormal tokens. Specifically, $\{M_{i,j}\}_{j=1}^L$ is utilized to identify anomalies within the latent token features $z_{i,j}^\text{S}$ (\textit{i.e.}, `\textbf{M}' operation in the Fig \ref{fig:main}). We introduce trainable mask tokens, $z_{mask}$, to replace these abnormal tokens in the latent feature space. Consequently, given the decoder $\mathbf{R}$, the restored image can be obtained as follows:
\begin{align}
&\{\{y_{i,j}\}_{j=1}^L\}_{i=1}^B=\mathbf{R}(\{\{\tilde{z}_{i,j}^\text{S}\}_{j=1}^L\}_{i=1}^B), \\
&
\tilde{z}_{i,j}^\text{S} =
\begin{cases}
z_{i,j}^\text{S}, & \textit{if } M_{i,j}=0 \\
z_{mask}, & \textit{if } M_{i,j}=1.
\end{cases}
\end{align}

To improve the quality of generated abnormal regions while preserving anatomical details, we introduce a weighted mean squared error (MSE) loss function:
\begin{align}
\mathcal{L}_\text{recon} = \frac{1}{L}\sum_{j=1}^L w_j \| y_{i,j} - x_{i,j} \|^2,
\end{align}
where $w_j$ is the weighted value to emphasize anomalies. 

\subsection{Overall Training Objective}
Our training methodology synergistically optimizes the aforementioned losses to enhance fine-grained anatomical discrimination in radiographic representation. The overall training objective is delineated as follows:
\begin{align}
\mathcal{L}(x_{i,j},x^\prime_{i,j}) = \mathcal{L}_\text{cst}(x_{i,j},x^\prime_{i,j}) + \mathcal{L}_\text{recon}(x_{i,j},x^\prime_{i,j})
\end{align}

\section{EXPERIMENT}
\subsection{Implementation Details} 

\textbf{\textit{Pre-training Dataset.}}
In this paper, we assemble a dataset of 811,170 Chest X-ray (CXR) images for pre-training, which includes 81,117 normal CXR images sourced from MIMIC-CXR-JPG \cite{mimic} and $81,117 \times 9$ synthetic abnormal images. Each normal CXR image is employed to generate nine synthetic abnormal images.

\textbf{\textit{Fine-tuning Datasets.}}
We evaluate our model on three CXR datasets: (i) \textbf{\textit{Child}CXRs} \cite{cell}, (ii) \textbf{NIH} \cite{Chestx-ray8}, (iii) \textbf{CXP} \cite{chexpert}, and (iv) \textbf{SIIM-ARC} \footnote{https://www.kaggle.com/c/siim-acr-pneumothorax-segmentation \label{kaggle:siim}}, using the official data partitions.
Comprehensive dataset details are deferred to Appendix C.

\textbf{\textit{Training Settings.}}
We follow the DINO \cite{dino} training paradigm (\textit{e.g.}, optimizer, learning rate, and weight decay schedule), utilizing an input image size of $224 \times 224$.
For model evaluation, we use the \textit{pre-training (on source data)} $\rightarrow$ \textit{fine-tuning (on target data)} protocol and employ two settings: transfer learning and anomaly detection. We implement our pre-training model using PyTorch and distribute the training across 8 NVIDIA A6000 GPUs, running for 800 epochs with a batch size of 64 for 6 days. 
We describe additional details on the pre-training and fine-tuning process in Appendix D.

\begin{table}[t]
  \centering
  \resizebox{\columnwidth}{!}{%
    \begin{tabular}{cccccc|ccc}
    \toprule
    \# & Aug. & \textit{Proto.} & \(\mathcal{L}_{\text{cst}}^\text{cate.}\) & \(\mathcal{L}_{\text{cst}}^\text{stru.}\) & \(\mathcal{L}_{\text{recon}}\) & AUC & ACC & F1 \\ \midrule
    1 & SLM          & \CheckmarkBold & \CheckmarkBold & \CheckmarkBold & \CheckmarkBold & \(\mathbf{89.4}\) & \(\mathbf{83.8}\) & \(\mathbf{84.5}\) \\
    2 & AnatPaste    & \CheckmarkBold & \CheckmarkBold & \CheckmarkBold & \CheckmarkBold & 88.9 & 83.2 & 83.6 \\
    3 & SLM          & \XSolidBrush   & \CheckmarkBold & \CheckmarkBold & \CheckmarkBold & 83.1 & 75.8 & 79.7 \\
    4 & SLM          & \CheckmarkBold & \XSolidBrush   & \XSolidBrush   & \CheckmarkBold & 76.9 & 67.2 & 75.9 \\
    5 & SLM          & \CheckmarkBold & \XSolidBrush   & \CheckmarkBold & \CheckmarkBold & 87.7 & 79.8 & 78.4\\
    6 & SLM          & \CheckmarkBold & \CheckmarkBold & \XSolidBrush   & \CheckmarkBold & 84.5 & 74.9 & 76.9\\ 
    7 & SLM          & \CheckmarkBold & \CheckmarkBold & \CheckmarkBold & \XSolidBrush   & 86.2 & 80.8 & 81.5 \\ 
    \bottomrule
    \end{tabular}%
  }
  \caption{\textbf{Impact of different components.} AUC scores $\uparrow$ (\%), ACC scores $\uparrow$ (\%), F1 scores $\uparrow$ (\%) on \textit{Child}CXR. }
  \label{tab:Ablation}
\end{table}

\begin{figure}[t]
  \centering
  \includegraphics[width=1\linewidth]{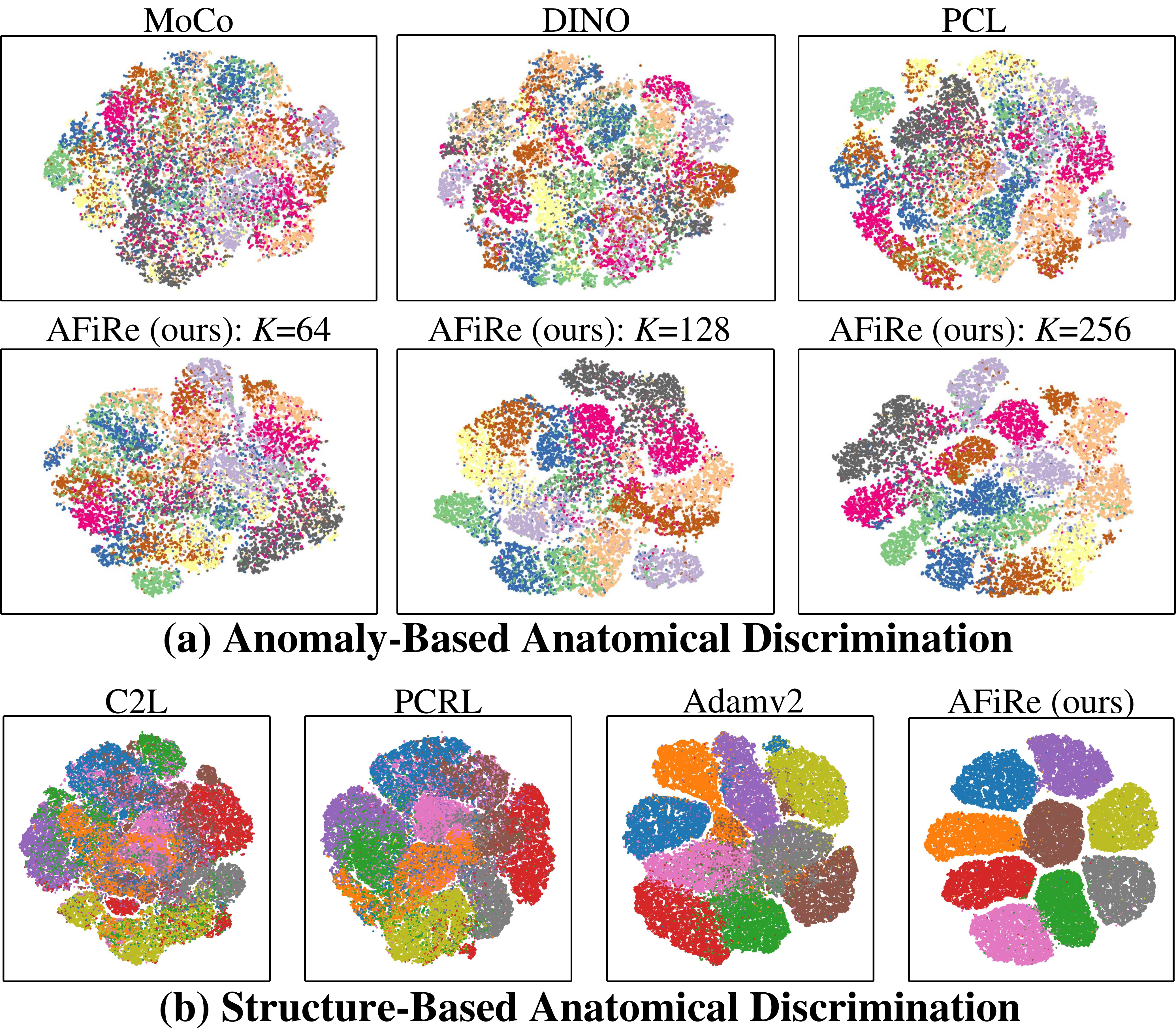}
  \caption{\textbf{T-SNE visualizations of the learned representation.} Different colors represent various anomalies (disease classes) in (a) and different image locations in (b).}
  \label{fig:cluster}
\end{figure}

\begin{table*}[t]
\centering
\resizebox{1\textwidth}{!}{%
\begin{tabular}{l|cccc|cccc|cccc}
        \toprule
        \multirow{2}{*}{Method} & \multicolumn{4}{c|}{NIH} & \multicolumn{4}{c|}{CXP} & \multicolumn{4}{c}{SIIM-ACR}\\
                                & 1\%    & 10\%   & 100\%  & 5-C    & 1\%    & 10\%  & 100\%  & 5-C     & 1\%    & 10\%  & 100\%  & 5-C\\\midrule
        ViTB \cite{vit}         & 57.1   & 67.5   & 80.1   & 79.9   & 75.1   & 84.0  & 87.4   & 85.3    & 34.9   & 56.9  & 73.2   & 72.3   \\
        MoCov2 \cite{mocoV2}    & 59.2   & 68.9   & 80.8   & 80.6   & 76.2   & 84.3  & 87.0   & 86.3    & 30.7   & 53.7  & 74.8   & 72.5 \\
        DINO \cite{dino}        & 60.6   & 69.6   & 81.0   & 80.3   & 76.9   & 84.9  & 87.6   & 86.1    & 35.8   & 57.4  & 77.4   & 74.8   \\
        MAE \cite{MAE}          & 60.5   & 70.1   & 80.9   & 81.7   & 77.7   & 85.2  & 88.0   & 86.5    & 41.2   & 60.9  & 79.3   & 79.1   \\
        C2L \cite{C2L}          & 61.7   & 73.1   & 81.4   & 82.1   & 77.6   & 85.4  & 89.3   & 86.1    & 38.9   & 63.7  & 74.7   & 71.4   \\
        MG \cite{MG}            & -      & -      & 80.8   & -      &  -     &  -   & 87.5    & -       & -      & -     & -      & -    \\
        TransVW \cite{TransVW}  & 61.2   & 69.7   & 81.2   & 81.9   & 78.2   & 84.5  & 88.2   & 86.7    & 42.7   & 66.9  & 76.7   & 72.3 \\
        DiRA \cite{Dira}        & 62.6   & 74.9   & 82.7   & 83.0   & 78.4   & 85.2  & 87.6   & 87.2    & 55.3   & 68.6  & 83.9   & 77.8   \\
        PCRL \cite{PCRL}        & 62.9   & 75.8   & 83.0   & 83.3   & 78.1   & 85.5  & 87.9   & 87.3    & 62.1   & 72.8  & 81.3   & 80.4 \\
        Adamv1 \cite{Adamv1}    & 60.5   & 71.6   & 81.2   & 82.0   & 77.9   & 84.9  & 87.7   & 86.2    & 48.7   & 67.2  & 72.8   & 69.9   \\
        Adamv2 \cite{Adamv2}    & 61.6   & 72.3   & 82.1   & 84.1   & 78.2   & 85.6  & 88.6   & 87.5    & 54.2   & 71.9  & 80.2   & 76.3   \\\midrule
        AFiRe (ours)            & \textbf{63.2}& \textbf{76.1}& \textbf{83.2}& \textbf{84.5}& \textbf{78.8}&\textbf{86.3}&\textbf{89.6}& \textbf{89.0} & \textbf{68.4}&\textbf{76.3}&\textbf{92.4}& \textbf{87.7}\\\bottomrule
        \end{tabular}%
        }
\caption{\textbf{Transfer learning with different labeling ratios.} AUC $\uparrow$ (\%) for multi-label classification on NIH and CXP, and Dice $\uparrow$ (\%) for segmentation on SIIM-ACR are reported. The best results are bolded. `5-C' denotes the five-fold cross-validation.}
\label{tab:Transfer}
\end{table*}

\subsection{Ablation Study}
Table \ref{tab:Ablation} elucidates the impact of various components in the proposed AFiRe. We evaluate the linear probing performance on the \textit{Child}CXR dataset by reporting AUC, ACC, and F1 scores. The components analyzed include different augmentation techniques (SLM vs. AnatPaste), the integration of the spacial-aware prototypes (\textit{Proto.}), and different pre-training proxy tasks such as category-consistency contrastive learning ($\mathcal{L}_{\text{cst}}^\text{cate.}$), structure-consistency contrastive learning ($\mathcal{L}_{\text{cst}}^\text{stru.}$), and anomaly-removal pixel restoration ($\mathcal{L}_{\text{recon}}$). The results reveal that our proposed model achieves peak performance with an AUC of $89.4\%$, ACC of $83.8\%$, and F1 of $84.5\%$ when all components are active (row 1). 

Substituting SLM with AnatPaste (row 2) leads to a slight performance drop across all metrics, indicating SLM's superior benefits and simplicity in generating pseudo lesions due to random $\eta$ and $\delta(w,h)$ selection. The suboptimal results in row 3 underline the necessity of spatial-aware prototypes for fine-grained anatomical discrimination. Rows 4, 5, and 6 highlight the importance of the proposed contrastive loss, as the removal of any one of them results in a significant performance decrease. Row 7 shows a notable performance decrease when the restoration task is omitted, underscoring the importance of pixel-level information for radiographic image analysis.

We further analyze the impact of other components in Appendix E, such as the contrastive protocol in Task \uppercase\expandafter{\romannumeral1}, the ablation of $z_{mask}$, and the integration of $I(Q_{jb}^{\text{S},-}, Q_{ji}^{\text{S},-})$.

\begin{figure}[ht]
  \centering
  \includegraphics[width=1\linewidth]{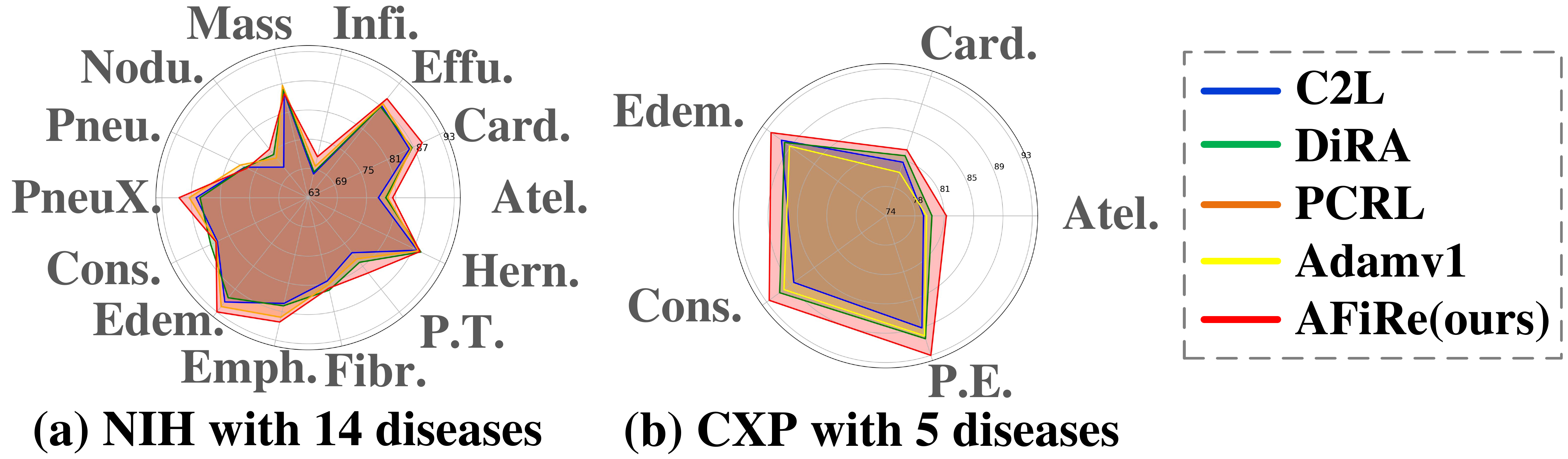}
  \caption{\textbf{Comparison of AUC scores $\uparrow$ (\%) for each disease} on the NIH dataset and CXP dataset using five-fold cross-validation.}
  \label{fig:compare}
\end{figure}

\subsection{Comparisons with State of the Arts}
\textbf{\textit{AFiRe achieves robust anatomical discrimination.}}
We use t-SNE \cite{t-SNE} to visualize the learned representations of AFiRe and other comparable methods, focusing on (i) anomaly-based and (ii) structure-based anatomical discrimination
For experiment (i), we leverage pre-trained parameters from MoCo \cite{mocoV2}, DINO \cite{dino}, PCL \cite{pcl}, and our AFiRe under varying cluster numbers ($K$) to extract radiographic representations. 
As shown in Fig. \ref{fig:cluster} (a), AFiRe achieves superior clustering performance for anomalies across 12 disease classes compared to other methods, attributed to its token-wise representation contrast enhanced by spatial-aware prototypes. The distinct separation of clusters becomes increasingly evident with higher values of $K$, \textit{indicating the effectiveness of AFiRe in discriminating complex anatomical features in real-world diseases.} This observation aligns with DINO's finding that a larger output dimensionality enhances performance. 
The detailed settings of this experiment are shown in Appendix F.
For experiment (ii), we adopt the settings of Adamv2 to annotate different image localizations (9 patches) as distinct anatomical landmarks. AFiRe is compared with three SOTA medical SSL methods: C2L \cite{C2L}, PCRL \cite{PCRL}, and Adamv2 \cite{Adamv2}. As shown in Fig. \ref{fig:cluster}(b), AFiRe achieves more cohesive feature clustering in structural anatomical discrimination. This result highlights AFiRe's ability to \textit{effectively differentiate various fine-grained anatomical structures.}

\textbf{\textit{AFiRe demonstrates effective generalization.}}
We evaluate the generalization ability of AFiRe by comparing it with 11 pre-training methods, reporting AUC scores for multi-label classification and Dice scores for segmentation across different labeling ratios (1\%, 10\%, and 100\%) on three downstream datasets. To ensure robustness, we further conduct five-fold cross-validation for these experiments.

\begin{table}[t]
\centering
        \resizebox{\columnwidth}{!}{%
        \begin{tabular}{l|ccc|ccc|ccc}
        \toprule
        \multirow{2}{*}{Model}     & \multicolumn{3}{c|}{\textit{Child}CXRs}              & \multicolumn{3}{c|}{CXP}                             & \multicolumn{3}{c}{NIH}\\
                                   & ACC             & AUC             & F1              & ACC             & AUC             & F1              & ACC             & AUC             & F1\\\midrule
        MemAE \cite{MemAE}         & $56.5$          & $77.8$          & $82.6$          & $55.6$          & $54.3$          & $53.3$          & $53.3$          & $54.0$          & $50.6$\\
        CutPast \cite{cutpaste}    & $64.0$          & $73.6$          & $72.3$          & $62.7$          & $65.5$          & $60.0$          & $-$             & $-$             & $-$           \\
        PANDA \cite{panda}         & $65.4$          & $65.7$          & $66.3$          & $66.4$          & $68.6$          & $65.3$          & $55.6$          & $57.4$          & $52.9$\\
        M-KD \cite{M-KD}           & $69.1$          & $74.1$          & $62.3$          & $66.0$          & $69.8$          & $63.6$          & $59.5$          & $61.7$          & $54.7$\\
        SALAD \cite{salad}         & $75.9$          & $82.6$          & $82.1$          & $-$             & $-$             & $-$             & $-$             & $-$             & $-$           \\
        f-AnoGAN \cite{f-AnoGAN}   & $74.0$          & $75.5$          & $81.0$          & $63.7$          & $65.8$          & $59.4$          & $62.6$          & $66.5$          & $58.6$\\
        SQUID \cite{SQUID}         & $80.3$          & $87.6$          & $84.7$          & $71.9$          & $78.1$          & $\mathbf{75.9}$ & $63.7$          & $66.9$          & $59.4$\\
        AnatPaste \cite{Anatpaste} & $83.0$          & $\mathbf{91.4}$ & $86.8$          & $70.4$          & $79.2$          & $73.5$          & $63.8$          & $67.4$          & $59.6$\\\midrule
        AFiRe (ours)                & $\mathbf{83.9}$ & $90.8$          & $\mathbf{87.6}$ & $\mathbf{72.4}$ &$\mathbf{79.9}$  & ${73.8}$        & $\mathbf{66.8}$ & $\mathbf{68.2}$ & $\mathbf{60.4}$ \\\bottomrule
        \end{tabular}%
        }
        \caption{\textbf{Classification results in anomaly detection.} Models are evaluated on ChildCXRs and NIH using official data split and the same settings as SQUID on CXP. ACC scores $\uparrow$ (\%), AUC scores $\uparrow$ (\%), F1 scores $\uparrow$ (\%) are reported.}
        \label{tab:Anomaly detection}
\end{table}

The results of the comparison are presented in Table \ref{tab:Transfer}. AFiRe was validated on NIH, CXP, and SIIM-ACR against: (i) a supervised baseline (ViTB pre-trained on ImageNet), demonstrating a significant improvement in radiographic image analysis ($+4.6\%$ AUC on NIH, $+3.7\%$ AUC on CXP and $+19.2\%$ Dice on SIIM-ACR) when using our SSL tasks; (ii) various SSL methods pre-trained on ImageNet, including MoCoV2, DINO, and MAE, with AFiRe achieving statistically significant improvements ($p<0.01$); and (iii) SSL methods specifically pre-trained on radiographic images, e.g., C2L, TransVW, MG, DiRA, PCRL, Adamv1, and Adamv2. AFiRe consistently outperforms these radiography-specific methods, especially at lower labeling ratios, highlighting its generalization in data-scarce scenarios.
Moreover, Fig. \ref{fig:compare} provides a detailed comparison of the AUC scores for each disease across the NIH and CXP datasets. In medical imaging, Cardiomegaly typically maintains well-defined borders despite an enlarged silhouette, while Edema and Emphysema present with varying degrees of increased or decreased density and often exhibit blurred or ill-defined borders. Our model excels in detecting these diseases, outperforming recent state-of-the-art methods with top AUC scores of $89.4\%$, $93.4\%$, and $89.6\%$ on NIH, respectively. \textit{These results illustrate that AFiRe can generalized to real disease types by pre-training with the SLM-augmented synthetic images.}

For the models in (i) and (ii), we re-implement them on NIH, CXP, and SIIM-ACR using publicly available ViT-B pre-trained parameters and the same data splits. For methods in (iii), we report original AUC scores when available; otherwise, we re-implement them.
More details and statistical significance analysis are shown in Appendix G.

\begin{figure}[t]
  \centering
  \includegraphics[width=1\linewidth]{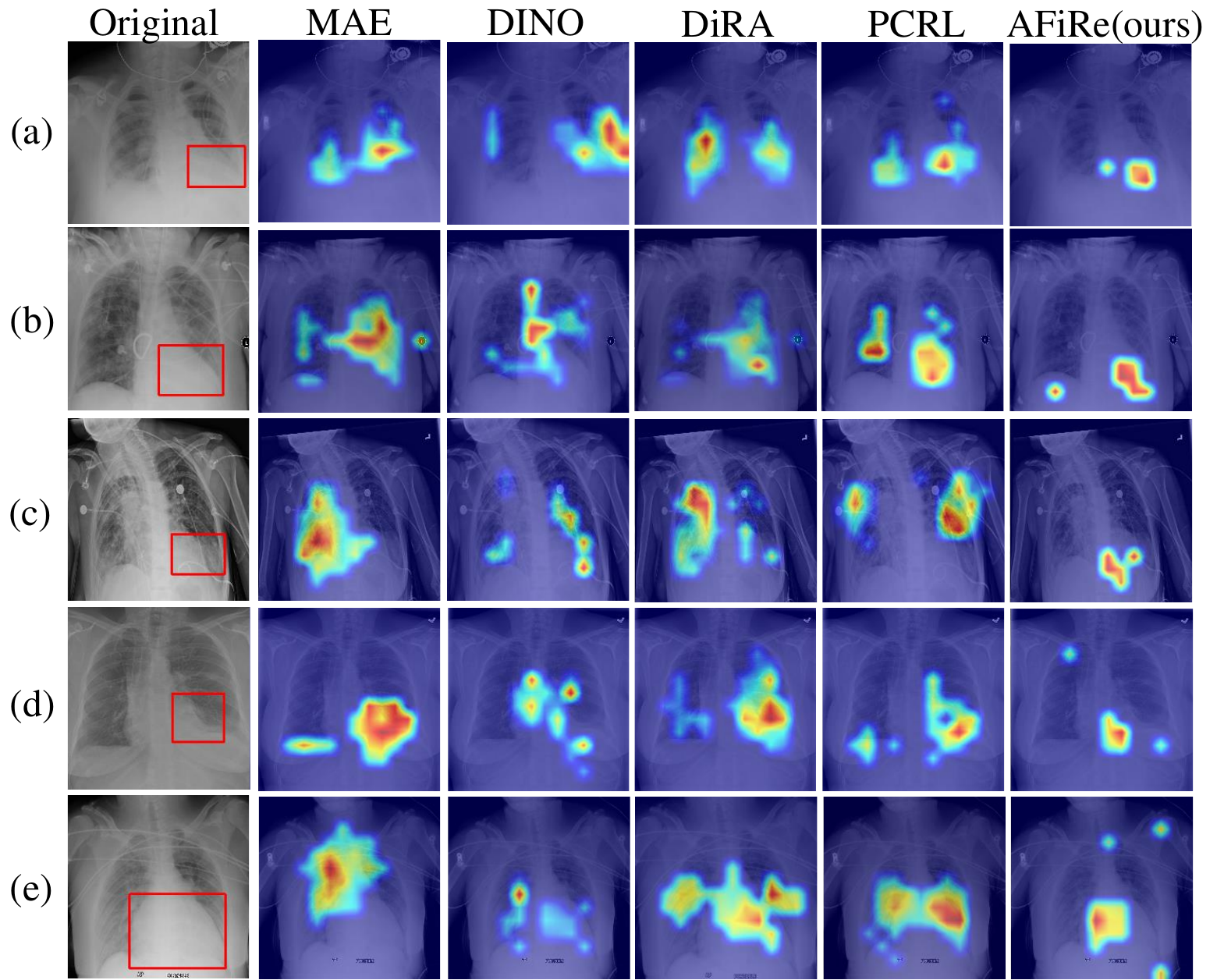}
  \caption{{Grad-CAM visualizations on NIH dataset.} The lesion regions are annotated by red bounding boxes.}
  \label{fig:cam}
\end{figure}

\textbf{\textit{AFiRe integrates fine-grained information.}}
We evaluate AFiRe's effectiveness in learning fine-grained representations through anomaly detection experiments.
In this experiment, both the pre-trained encoder and decoder are unsupervisedly fine-tuned using Tasks \uppercase\expandafter{\romannumeral1} and \uppercase\expandafter{\romannumeral2} on the target training set. Table \ref{tab:Anomaly detection} presents the results across three datasets, \textit{Child}CXRs, CXP, and NIH, against: (i) recent unsupervised anomaly detection methods for natural images (e.g., MemAE, CutPast, PANDA, M-KD), where AFiRe shows significant improvements (average AUC +15\%, +12\%, and +10\% on \textit{Child}CXRs, CXP, and NIH, respectively), and (ii) state-of-the-art unsupervised methods for medical images (e.g., SALAD, f-AnoGAN, SQUID, AnatPaste), where AFiRe maintains a significant improvement ($p<0.05$) across all tasks, demonstrating a robust balance between precision and recall. 
We also present Grad-CAM visualizations to compare the performance of various models, including MAE, DINO, DiRA, PCRL, and the proposed AFiRe, in highlighting lesion regions, as shown in Fig. \ref{fig:cam}. Compared to the other models, AFiRe demonstrates more precise and focused lesion regions. Especially, in row (e), despite missing some parts of the lesion, AFiRe shows a clear advantage in precision and minimizing false positives, highlighting the main lesion area more effectively than the other models. \textit{This indicates that AFiRe can better capture and pay attention to specific local anatomical details, effectively distinguishing lesions from surrounding tissues.} 
More details and statistical significance analysis are shown in Appendix H.

\section{Conclusion}
 We propose an anatomy-driven self-supervised framework, AFiRe, to incorporate fine-grained anatomical discriminative representations. We synergistically perform this framework with Token-wise anatomy-guided contrastive learning and Pixel-level anomaly-removal restoration.
 Experimental results on downstream radiographic image diagnosis tasks confirm the robust generalization. Our proposed model exhibits outstanding performance in multi-label classification and segmentation. It also provides effective anomaly detection capabilities. However, it tends to prioritize primary abnormalities, potentially overlooking larger affected regions. Our future research is going to improve the model's capability for more accurate detection of lesion regions.

\section{Acknowledgments}
AAAI is especially grateful to Peter Patel Schneider for his work in implementing the original aaai.sty file, liberally using the ideas of other style hackers, including Barbara Beeton. We also acknowledge with thanks the work of George Ferguson for his guide to using the style and BibTeX files --- which has been incorporated into this document --- and Hans Guesgen, who provided several timely modifications, as well as the many others who have, from time to time, sent in suggestions on improvements to the AAAI style. We are especially grateful to Francisco Cruz, Marc Pujol-Gonzalez, and Mico Loretan for the improvements to the Bib\TeX{} and \LaTeX{} files made in 2020.

\bibliography{aaai25}

\begin{thebibliography}{43}
\providecommand{\natexlab}[1]{#1}

\bibitem[{Agu et~al.(2021)Agu, Wu, Chao, Lourentzou, Sharma, Moradi, Yan, and Hendler}]{AnaXNet}
Agu, N.~N.; Wu, J.~T.; Chao, H.; Lourentzou, I.; Sharma, A.; Moradi, M.; Yan, P.; and Hendler, J. 2021.
\newblock AnaXNet: anatomy aware multi-label finding classification in chest X-ray.
\newblock In \emph{Medical Image Computing and Computer Assisted Intervention--MICCAI 2021: 24th International Conference, Strasbourg, France, September 27--October 1, 2021, Proceedings, Part V 24}, 804--813. Springer.

\bibitem[{Azizi et~al.(2021)Azizi, Mustafa, Ryan, Beaver, Freyberg, Deaton, Loh, Karthikesalingam, Kornblith, Chen et~al.}]{MICLe_contrast_CXR}
Azizi, S.; Mustafa, B.; Ryan, F.; Beaver, Z.; Freyberg, J.; Deaton, J.; Loh, A.; Karthikesalingam, A.; Kornblith, S.; Chen, T.; et~al. 2021.
\newblock Big self-supervised models advance medical image classification.
\newblock In \emph{Proceedings of the IEEE/CVF international conference on computer vision}, 3478--3488.

\bibitem[{Caron et~al.(2020)Caron, Misra, Mairal, Goyal, Bojanowski, and Joulin}]{swav}
Caron, M.; Misra, I.; Mairal, J.; Goyal, P.; Bojanowski, P.; and Joulin, A. 2020.
\newblock Unsupervised learning of visual features by contrasting cluster assignments.
\newblock \emph{Advances in neural information processing systems}, 33: 9912--9924.

\bibitem[{Caron et~al.(2021)Caron, Touvron, Misra, J{\'e}gou, Mairal, Bojanowski, and Joulin}]{dino}
Caron, M.; Touvron, H.; Misra, I.; J{\'e}gou, H.; Mairal, J.; Bojanowski, P.; and Joulin, A. 2021.
\newblock Emerging properties in self-supervised vision transformers.
\newblock In \emph{Proceedings of the IEEE/CVF international conference on computer vision}, 9650--9660.

\bibitem[{Chen et~al.(2023)Chen, Wang, Wang, Li, Fang, Li, Bai, Peng, Meng, and Wang}]{discrimination_pami}
Chen, H.; Wang, R.; Wang, X.; Li, J.; Fang, Q.; Li, H.; Bai, J.; Peng, Q.; Meng, D.; and Wang, L. 2023.
\newblock Unsupervised local discrimination for medical images.
\newblock \emph{IEEE Transactions on Pattern Analysis and Machine Intelligence}.

\bibitem[{Chen et~al.(2020{\natexlab{a}})Chen, Kornblith, Norouzi, and Hinton}]{SimCLR}
Chen, T.; Kornblith, S.; Norouzi, M.; and Hinton, G. 2020{\natexlab{a}}.
\newblock A simple framework for contrastive learning of visual representations.
\newblock In \emph{International conference on machine learning}, 1597--1607. PMLR.

\bibitem[{Chen et~al.(2020{\natexlab{b}})Chen, Fan, Girshick, and He}]{mocoV2}
Chen, X.; Fan, H.; Girshick, R.; and He, K. 2020{\natexlab{b}}.
\newblock Improved baselines with momentum contrastive learning.
\newblock \emph{arXiv preprint arXiv:2003.04297}.

\bibitem[{Cuturi(2013)}]{Sinkhorn}
Cuturi, M. 2013.
\newblock Sinkhorn distances: Lightspeed computation of optimal transport.
\newblock \emph{Advances in neural information processing systems}, 26.

\bibitem[{Dosovitskiy et~al.(2021)Dosovitskiy, Beyer, Kolesnikov, Weissenborn, Zhai, Unterthiner, Dehghani, Minderer, Heigold, and Gelly}]{vit}
Dosovitskiy, A.; Beyer, L.; Kolesnikov, A.; Weissenborn, D.; Zhai, X.; Unterthiner, T.; Dehghani, M.; Minderer, M.; Heigold, G.; and Gelly, S.~a. 2021.
\newblock An Image is Worth 16x16 Words: Transformers for Image Recognition at Scale.
\newblock In \emph{International Conference on Learning Representations}.

\bibitem[{Gong et~al.(2019)Gong, Liu, Le, Saha, Mansour, Venkatesh, and Hengel}]{MemAE}
Gong, D.; Liu, L.; Le, V.; Saha, B.; Mansour, M.~R.; Venkatesh, S.; and Hengel, A. v.~d. 2019.
\newblock Memorizing normality to detect anomaly: Memory-augmented deep autoencoder for unsupervised anomaly detection.
\newblock In \emph{Proceedings of the IEEE/CVF International Conference on Computer Vision}, 1705--1714.

\bibitem[{Haghighi et~al.(2022)Haghighi, Taher, Gotway, and Liang}]{Dira}
Haghighi, F.; Taher, M. R.~H.; Gotway, M.~B.; and Liang, J. 2022.
\newblock Dira: Discriminative, restorative, and adversarial learning for self-supervised medical image analysis.
\newblock In \emph{Proceedings of the IEEE/CVF Conference on Computer Vision and Pattern Recognition}, 20824--20834.

\bibitem[{Haghighi et~al.(2024)Haghighi, Taher, Gotway, and Liang}]{Dirav2}
Haghighi, F.; Taher, M. R.~H.; Gotway, M.~B.; and Liang, J. 2024.
\newblock Self-supervised learning for medical image analysis: Discriminative, restorative, or adversarial?
\newblock \emph{Medical Image Analysis}, 94: 103086.

\bibitem[{Haghighi et~al.(2021)Haghighi, Taher, Zhou, Gotway, and Liang}]{TransVW}
Haghighi, F.; Taher, M. R.~H.; Zhou, Z.; Gotway, M.~B.; and Liang, J. 2021.
\newblock Transferable visual words: Exploiting the semantics of anatomical patterns for self-supervised learning.
\newblock \emph{IEEE transactions on medical imaging}, 40(10): 2857--2868.

\bibitem[{He et~al.(2022)He, Chen, Xie, Li, Doll{\'a}r, and Girshick}]{MAE}
He, K.; Chen, X.; Xie, S.; Li, Y.; Doll{\'a}r, P.; and Girshick, R. 2022.
\newblock Masked autoencoders are scalable vision learners.
\newblock In \emph{Proceedings of the IEEE/CVF conference on computer vision and pattern recognition}, 16000--16009.

\bibitem[{He et~al.(2020)He, Fan, Wu, Xie, and Girshick}]{moco}
He, K.; Fan, H.; Wu, Y.; Xie, S.; and Girshick, R. 2020.
\newblock Momentum contrast for unsupervised visual representation learning.
\newblock In \emph{Proceedings of the IEEE/CVF conference on computer vision and pattern recognition}, 9729--9738.

\bibitem[{Hosseinzadeh~Taher, Gotway, and Liang(2023)}]{Adamv1}
Hosseinzadeh~Taher, M.~R.; Gotway, M.~B.; and Liang, J. 2023.
\newblock Towards foundation models learned from anatomy in medical imaging via self-supervision.
\newblock In \emph{MICCAI Workshop on Domain Adaptation and Representation Transfer}, 94--104. Springer.

\bibitem[{Huang et~al.(2021)Huang, Lin, Cheng, Lyu, and Tang}]{pre_roi}
Huang, Y.; Lin, L.; Cheng, P.; Lyu, J.; and Tang, X. 2021.
\newblock Lesion-based contrastive learning for diabetic retinopathy grading from fundus images.
\newblock In \emph{Medical Image Computing and Computer Assisted Intervention--MICCAI 2021: 24th International Conference, Strasbourg, France, September 27--October 1, 2021, Proceedings, Part II 24}, 113--123. Springer.

\bibitem[{Irvin et~al.(2019)Irvin, Rajpurkar, Ko, Yu, Ciurea-Ilcus, Chute, Marklund, Haghgoo, Ball, Shpanskaya et~al.}]{chexpert}
Irvin, J.; Rajpurkar, P.; Ko, M.; Yu, Y.; Ciurea-Ilcus, S.; Chute, C.; Marklund, H.; Haghgoo, B.; Ball, R.; Shpanskaya, K.; et~al. 2019.
\newblock Chexpert: A large chest radiograph dataset with uncertainty labels and expert comparison.
\newblock In \emph{Proceedings of the AAAI conference on artificial intelligence}, volume 33(01), 590--597.

\bibitem[{Johnson et~al.(2019)Johnson, Pollard, Greenbaum, Lungren, Deng, Peng, Lu, Mark, Berkowitz, and Horng}]{mimic}
Johnson, A.~E.; Pollard, T.~J.; Greenbaum, N.~R.; Lungren, M.~P.; Deng, C.-y.; Peng, Y.; Lu, Z.; Mark, R.~G.; Berkowitz, S.~J.; and Horng, S. 2019.
\newblock MIMIC-CXR-JPG, a large publicly available database of labeled chest radiographs.
\newblock \emph{arXiv preprint arXiv:1901.07042}.

\bibitem[{Kermany et~al.(2018)Kermany, Goldbaum, Cai, Valentim, Liang, Baxter, McKeown, Yang, Wu, Yan, Dong, Prasadha, Pei, Ting, Zhu, Li, Hewett, Dong, Ziyar, Shi, Zhang, Zheng, Hou, Shi, Fu, Duan, Huu, Wen, Zhang, Zhang, Li, Wang, Singer, Sun, Xu, Tafreshi, Lewis, Xia, and Zhang}]{cell}
Kermany, D.~S.; Goldbaum, M.; Cai, W.; Valentim, C.~C.; Liang, H.; Baxter, S.~L.; McKeown, A.; Yang, G.; Wu, X.; Yan, F.; Dong, J.; Prasadha, M.~K.; Pei, J.; Ting, M.~Y.; Zhu, J.; Li, C.; Hewett, S.; Dong, J.; Ziyar, I.; Shi, A.; Zhang, R.; Zheng, L.; Hou, R.; Shi, W.; Fu, X.; Duan, Y.; Huu, V.~A.; Wen, C.; Zhang, E.~D.; Zhang, C.~L.; Li, O.; Wang, X.; Singer, M.~A.; Sun, X.; Xu, J.; Tafreshi, A.; Lewis, M.~A.; Xia, H.; and Zhang, K. 2018.
\newblock Identifying Medical Diagnoses and Treatable Diseases by Image-Based Deep Learning.
\newblock \emph{Cell}, 172(5): 1122--1131.e9.

\bibitem[{Li et~al.(2021)Li, Sohn, Yoon, and Pfister}]{cutpaste}
Li, C.-L.; Sohn, K.; Yoon, J.; and Pfister, T. 2021.
\newblock Cutpaste: Self-supervised learning for anomaly detection and localization.
\newblock In \emph{Proceedings of the IEEE/CVF conference on computer vision and pattern recognition}, 9664--9674.

\bibitem[{Li et~al.(2020)Li, Zhou, Xiong, and Hoi}]{pcl}
Li, J.; Zhou, P.; Xiong, C.; and Hoi, S.~C. 2020.
\newblock Prototypical contrastive learning of unsupervised representations.
\newblock \emph{arXiv preprint arXiv:2005.04966}.

\bibitem[{Li et~al.(2024)Li, Yang, Ren, Nie, Gao, Tan, and Li}]{mlip}
Li, Z.; Yang, L.~T.; Ren, B.; Nie, X.; Gao, Z.; Tan, C.; and Li, S.~Z. 2024.
\newblock MLIP: Enhancing Medical Visual Representation with Divergence Encoder and Knowledge-guided Contrastive Learning.
\newblock \emph{arXiv preprint arXiv:2402.02045}.

\bibitem[{Misra and Maaten(2020)}]{Contrastive-pretext-invariant}
Misra, I.; and Maaten, L. v.~d. 2020.
\newblock Self-supervised learning of pretext-invariant representations.
\newblock In \emph{Proceedings of the IEEE/CVF conference on computer vision and pattern recognition}, 6707--6717.

\bibitem[{Otsu et~al.(1975)}]{otsu}
Otsu, N.; et~al. 1975.
\newblock A threshold selection method from gray-level histograms.
\newblock \emph{Automatica}, 11(285-296): 23--27.

\bibitem[{Reiss et~al.(2021)Reiss, Cohen, Bergman, and Hoshen}]{panda}
Reiss, T.; Cohen, N.; Bergman, L.; and Hoshen, Y. 2021.
\newblock Panda: Adapting pretrained features for anomaly detection and segmentation.
\newblock In \emph{Proceedings of the IEEE/CVF Conference on Computer Vision and Pattern Recognition}, 2806--2814.

\bibitem[{Salehi et~al.(2021)Salehi, Sadjadi, Baselizadeh, Rohban, and Rabiee}]{M-KD}
Salehi, M.; Sadjadi, N.; Baselizadeh, S.; Rohban, M.~H.; and Rabiee, H.~R. 2021.
\newblock Multiresolution knowledge distillation for anomaly detection.
\newblock In \emph{Proceedings of the IEEE/CVF conference on computer vision and pattern recognition}, 14902--14912.

\bibitem[{Sato et~al.(2023)Sato, Suzuki, Wataya, Nishigaki, Kita, Yamagata, Tomiyama, and Kido}]{Anatpaste}
Sato, J.; Suzuki, Y.; Wataya, T.; Nishigaki, D.; Kita, K.; Yamagata, K.; Tomiyama, N.; and Kido, S. 2023.
\newblock Anatomy-aware self-supervised learning for anomaly detection in chest radiographs.
\newblock \emph{iScience}.

\bibitem[{Schlegl et~al.(2019)Schlegl, Seeb{\"o}ck, Waldstein, Langs, and Schmidt-Erfurth}]{f-AnoGAN}
Schlegl, T.; Seeb{\"o}ck, P.; Waldstein, S.~M.; Langs, G.; and Schmidt-Erfurth, U. 2019.
\newblock f-AnoGAN: Fast unsupervised anomaly detection with generative adversarial networks.
\newblock \emph{Medical image analysis}, 54: 30--44.

\bibitem[{Singh, Gorade, and Mishra(2024)}]{MLVICX}
Singh, A.; Gorade, V.; and Mishra, D. 2024.
\newblock MLVICX: Multi-Level Variance-Covariance Exploration for Chest X-ray Self-Supervised Representation Learning.
\newblock \emph{arXiv preprint arXiv:2403.11504}.

\bibitem[{Sowrirajan et~al.(2021)Sowrirajan, Yang, Ng, and Rajpurkar}]{MoCo-CXR}
Sowrirajan, H.; Yang, J.; Ng, A.~Y.; and Rajpurkar, P. 2021.
\newblock Moco pretraining improves representation and transferability of chest x-ray models.
\newblock In \emph{Medical Imaging with Deep Learning}, 728--744. PMLR.

\bibitem[{Taher, Gotway, and Liang(2024)}]{Adamv2}
Taher, M. R.~H.; Gotway, M.~B.; and Liang, J. 2024.
\newblock Representing Part-Whole Hierarchies in Foundation Models by Learning Localizability Composability and Decomposability from Anatomy via Self Supervision.
\newblock In \emph{Proceedings of the IEEE/CVF Conference on Computer Vision and Pattern Recognition}, 11269--11281.

\bibitem[{Van~der Maaten and Hinton(2008)}]{t-SNE}
Van~der Maaten, L.; and Hinton, G. 2008.
\newblock Visualizing data using t-SNE.
\newblock \emph{Journal of machine learning research}, 9(11).

\bibitem[{Vu et~al.(2021)Vu, Wang, Balachandar, Liu, Ng, and Rajpurkar}]{Medaug_contrast_CXR}
Vu, Y. N.~T.; Wang, R.; Balachandar, N.; Liu, C.; Ng, A.~Y.; and Rajpurkar, P. 2021.
\newblock Medaug: Contrastive learning leveraging patient metadata improves representations for chest x-ray interpretation.
\newblock In \emph{Machine Learning for Healthcare Conference}, 755--769. PMLR.

\bibitem[{Wang et~al.(2017)Wang, Peng, Lu, Lu, Bagheri, and Summers}]{Chestx-ray8}
Wang, X.; Peng, Y.; Lu, L.; Lu, Z.; Bagheri, M.; and Summers, R.~M. 2017.
\newblock Chestx-ray8: Hospital-scale chest x-ray database and benchmarks on weakly-supervised classification and localization of common thorax diseases.
\newblock In \emph{Proceedings of the IEEE conference on computer vision and pattern recognition}, 2097--2106.

\bibitem[{Wu et~al.(2018)Wu, Xiong, Yu, and Lin}]{Contrastive-instance-discrimination}
Wu, Z.; Xiong, Y.; Yu, S.~X.; and Lin, D. 2018.
\newblock Unsupervised feature learning via non-parametric instance discrimination.
\newblock In \emph{Proceedings of the IEEE conference on computer vision and pattern recognition}, 3733--3742.

\bibitem[{Xiang et~al.(2023)Xiang, Zhang, Lu, Yuille, Zhang, Cai, and Zhou}]{SQUID}
Xiang, T.; Zhang, Y.; Lu, Y.; Yuille, A.~L.; Zhang, C.; Cai, W.; and Zhou, Z. 2023.
\newblock SQUID: Deep Feature In-Painting for Unsupervised Anomaly Detection.
\newblock In \emph{Proceedings of the IEEE/CVF Conference on Computer Vision and Pattern Recognition}, 23890--23901.

\bibitem[{Yun et~al.(2022)Yun, Lee, Kim, and Shin}]{Patch-level}
Yun, S.; Lee, H.; Kim, J.; and Shin, J. 2022.
\newblock Patch-level representation learning for self-supervised vision transformers.
\newblock In \emph{Proceedings of the IEEE/CVF conference on computer vision and pattern recognition}, 8354--8363.

\bibitem[{Zhao et~al.(2021{\natexlab{a}})Zhao, Li, He, Ma, Fang, Li, and Zheng}]{salad}
Zhao, H.; Li, Y.; He, N.; Ma, K.; Fang, L.; Li, H.; and Zheng, Y. 2021{\natexlab{a}}.
\newblock Anomaly detection for medical images using self-supervised and translation-consistent features.
\newblock \emph{IEEE Transactions on Medical Imaging}, 40(12): 3641--3651.

\bibitem[{Zhao et~al.(2021{\natexlab{b}})Zhao, Cao, Yao, Nogues, Lu, Huang, Xiao, Yin, and Zhang}]{3D_anatomy}
Zhao, T.; Cao, K.; Yao, J.; Nogues, I.; Lu, L.; Huang, L.; Xiao, J.; Yin, Z.; and Zhang, L. 2021{\natexlab{b}}.
\newblock 3D graph anatomy geometry-integrated network for pancreatic mass segmentation, diagnosis, and quantitative patient management.
\newblock In \emph{Proceedings of the IEEE/CVF conference on computer vision and pattern recognition}, 13743--13752.

\bibitem[{Zhou et~al.(2021{\natexlab{a}})Zhou, Lu, Yang, Han, and Yu}]{PCRL}
Zhou, H.-Y.; Lu, C.; Yang, S.; Han, X.; and Yu, Y. 2021{\natexlab{a}}.
\newblock Preservational learning improves self-supervised medical image models by reconstructing diverse contexts.
\newblock In \emph{Proceedings of the IEEE/CVF International Conference on Computer Vision}, 3499--3509.

\bibitem[{Zhou et~al.(2020)Zhou, Yu, Bian, Hu, Ma, and Zheng}]{C2L}
Zhou, H.-Y.; Yu, S.; Bian, C.; Hu, Y.; Ma, K.; and Zheng, Y. 2020.
\newblock Comparing to learn: Surpassing imagenet pretraining on radiographs by comparing image representations.
\newblock In \emph{Medical Image Computing and Computer Assisted Intervention--MICCAI 2020: 23rd International Conference, Lima, Peru, October 4--8, 2020, Proceedings, Part I 23}, 398--407. Springer.

\bibitem[{Zhou et~al.(2021{\natexlab{b}})Zhou, Sodha, Pang, Gotway, and Liang}]{MG}
Zhou, Z.; Sodha, V.; Pang, J.; Gotway, M.~B.; and Liang, J. 2021{\natexlab{b}}.
\newblock Models genesis.
\newblock \emph{Medical image analysis}, 67: 101840.

\end{thebibliography}

\begin{figure}
  \centering
  \includegraphics[width=1\linewidth]{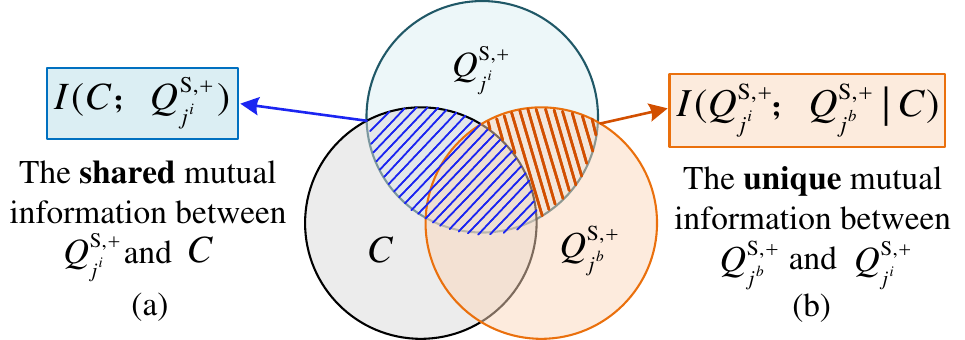}
  \caption{Visual depiction of the optimized mutual information among different normal token probabilities ($Q_{j^i}^{\text{S},+}$ and $Q_{j^b}^{\text{S},+}$) and the spatial-aware prototypes ($C$).}
  \label{fig:MI_visual}
\end{figure}

\section{A. Simplifying the Mutual Information}
Here, we present a proof for simplifying $I(Q_{i}^{\text{S},+}; C)$.
Assume that:
\begin{align}
&I(Q_{i}^{\text{S},+}; C) =\sum_{q_{i,j}^{\text{S},+}} \sum_{c_j} P(c_{j}, q_{i,j}^{\text{S},+}) \log \frac{P(c_j|q_{i,j}^{\text{S},+})}{P(c_j)},
  \label{eq:MI2} \\
&P(c_{j}, q_{i,j}^{\text{S},+})=P(c_{j}| q_{i,j}^{\text{S},+})P(q_{i,j}^{\text{S},+}), \label{eq:joint2}\\
&P(c_{j}| q_{i,j}^{\text{S},+})=\frac{\text{sim}(q_{i,j}^{\text{S},+}, c_j)}{\sum_{l=1}^L\text{sim}(q_{i,j}^{\text{S},+}, c_l)}.
\label{eq:conditional2}
\end{align}
Given that $Q_{i}^{\text{S},+}$ represents the probabilities of normal tokens and $C$ represents the probabilities of prototypes, the distributions $P(q_{i,j}^{\text{S},+})$ and $P(c_j)$ are fixed in each iteration. Consequently, the mutual information $I(Q_{i}^{\text{S},+}; C)$ depends solely on the conditional probability $P(c_{j}| q_{i,j}^{\text{S},+})$. Specifically, applying the properties of the logarithm function, we decompose:
\begin{align}
I(Q_{i}^{\text{S},+}; C) 
&=\sum_{q_{i,j}^{\text{S},+}} \sum_{c_j} P(q_{i,j}^{\text{S},+}, c_{j}) \log P(c_j|q_{i,j}^{\text{S},+}) \label{eq:term_1} \\
&\quad - \sum_{q_{i,j}^{\text{S},+}} \sum_{c_j} P(q_{i,j}^{\text{S},+}, c_{j}) \log P(c_j).
\label{eq:term_2}
\end{align}

Next, we apply Jensen's inequality to the Eq.\ref{eq:term_2}. By substituting Eq. \ref{eq:joint2} and ignoring the constant $P(q_{i,j}^{\text{S},+})$ and $P(c_j)$, we obtain:
\begin{align}
&\sum_{q_{i,j}^{\text{S},+}} \sum_{c_j} P(q_{i,j}^{\text{S},+}, c_{j}) \log P(c_j) \\
&\leq \log \sum_{q_{i,j}^{\text{S},+}} \sum_{c_j} P(c_{j}| q_{i,j}^{\text{S},+})P(q_{i,j}^{\text{S},+}) P(c_j), \\ 
&= \log \sum_{q_{i,j}^{\text{S},+}} \sum_{c_j} P(c_j|q_{i,j}^{\text{S},+}).
\label{eq:term_2.1}
\end{align}
Since the denominator in Eq. \ref{eq:conditional2} is constant under fixed $q_{i,j}^{\text{S},+}$ and $c_j$, $\sum_{c_j}\text{sim}(q_{i,j}^{\text{S},+}, c_j)$ is equivalent to $\sum_{l=1}^L\text{sim}(q_{i,j}^{\text{S},+}, c_l)$. Therefore, in the Eq. \ref{eq:term_2.1}:
\begin{align}
&\log \sum_{q_{i,j}^{\text{S},+}} \sum_{c_j} P(c_j|q_{i,j}^{\text{S},+}) \\
&=\log \sum_{q_{i,j}^{\text{S},+}} \frac{\sum_{c_j} \text{sim}(q_{i,j}^{\text{S},+}, c_j)}{\sum_{l=1}^L\text{sim}(q_{i,j}^{\text{S},+}, c_l)}, \\ 
&= \log \sum_{q_{i,j}^{\text{S},+}} 1, \\
&= \log L.
\label{eq:term_2.2}
\end{align}

Substituting Eq. \ref{eq:term_2.2} with Eq. \ref{eq:term_2}, we have: 
\begin{align}
& I(Q_{i}^{\text{S},+}; C) \\
&\geq \sum_{q_{i,j}^{\text{S},+}} \sum_{c_j} P(q_{i,j}^{\text{S},+}, c_{j}) \log P(c_j|q_{i,j}^{\text{S},+}) - \log L, \\
& = \mathbb{E}_{P(Q_{i}^{\text{S},+}, C)} \left[ \log \frac{\text{sim}(q_{i,j}^{\text{S},+}, c_j)}{\sum_{l=1}^L \text{sim}(q_{i,j}^{\text{S},+}, c_l)} \right]  - \log L, 
\end{align}
where $L$ is the constant about the length of image tokens. By ignoring the $\log L$, the Eq. \ref{eq:MI2} can be simplified as:
\begin{align}
& I(Q_{i}^{\text{S},+}; C) = \mathbb{E}_{P(Q_{i}^{\text{S},+}, C)} \left[ \log \frac{\text{sim}(q_{i,j}^{\text{S},+}, c_j)}{\sum_{l=1}^L \text{sim}(q_{i,j}^{\text{S},+}, c_l)} \right]. \label{eq:MI_result}
\end{align}

\section{B. Proof of $\mathcal{L}_{\text{cst}}^\text{cate.}$}
We aim to maximize the mutual information among the probabilities of normal-category tokens at the same anatomical structure, encompassing different image tokens ($Q_{j^i}^{\text{S},+} = \{q_{i,j}^{\text{S},+}\}_{i=1}^B$ and $Q_{j^b}^{\text{S},+} = \{q_{b,j}^{\text{S},+}\}_{b=1}^B$) and the same positional spatial-aware prototypes ($C= \{c_j\}_{j=1}^L$). Here, $q_{i,j}^{\text{S},+}$ and $q_{b,j}^{\text{S},+}$ are two different samples from the same range. Consequently, $I(C; Q_{j^b}^{\text{S},+})$ is obtained similarly to $I(C; Q_{j^i}^{\text{S},+})$. To this end, by considering all $j \in [1,L]$, we propose to optimize the joint mutual information function: $I(C, Q_{j^b}^{\text{S},+}; Q_{j^i}^{\text{S},+})$. As illustrated in Fig. \ref{fig:MI_visual}, $I(C, Q_{j^b}^{\text{S},+}; Q_{j^i}^{\text{S},+})$ consists of (i) the shared mutual information between $q_{i,j}^{\text{S},+}$ and $c_{j}$ ($I(C; Q_{j^i}^{\text{S},+})$, Fig. \ref{fig:MI_visual}(a)), and (ii) the unique mutual information between $q_{i,j}^{\text{S},+}$ and $q_{b,j}^{\text{S},+}$ ($I(Q_{j^i}^{\text{S},+};Q_{j^b}^{\text{S},+}|C)$, Fig. \ref{fig:MI_visual}(b)). This can be formulated as: 
\begin{equation}
\begin{aligned}
&I(C, Q_{j^b}^{\text{S},+}; Q_{j^i}^{\text{S},+}) \\
&= I(C; Q_{j^i}^{\text{S},+}) + I(Q_{j^i}^{\text{S},+};Q_{j^b}^{\text{S},+}|C), \\
&= \sum_{c_j, q_{i,j}^{\text{S},+},q_{b,j}^{\text{S},+}} P(c_{j}, q_{i,j}^{\text{S},+}, q_{b,j}^{\text{S},+}) \log \frac{P(q_{i,j}^{\text{S},+}|c_{j} q_{b,j}^{\text{S},+})}{P(q_{i,j}^{\text{S},+})}.
\label{eq:MI_cate2}
\end{aligned}
\end{equation}
By applying the properties of the logarithm function, we decompose Eq. \ref{eq:MI_cate2} to obtain:
\begin{equation}
\begin{aligned}
&I(C, Q_{j^b}^{\text{S},+}; Q_{j^i}^{\text{S},+}) \\
&= \sum_{c_j, q_{i,j}^{\text{S},+},q_{b,j}^{\text{S},+}} P(c_{j}, q_{i,j}^{\text{S},+}, q_{b,j}^{\text{S},+}) \log P(q_{i,j}^{\text{S},+}|c_{j} q_{b,j}^{\text{S},+}), \\
&\quad - \sum_{c_j, q_{i,j}^{\text{S},+},q_{b,j}^{\text{S},+}} P(c_{j}, q_{i,j}^{\text{S},+}, q_{b,j}^{\text{S},+}) \log P(q_{i,j}^{\text{S},+}).
\label{eq:MI_cate_decompose}
\end{aligned}
\end{equation}
Here, when considering all the $i\in [1,B^+]$ and $j\in [1,L]$, $I(C; Q_{j^i}^{\text{S},+}) \propto I(Q_{i}^{\text{S},+}; C)$, which is optimized by $\mathcal{L}_{\text{cst}}^\text{stru.}$. Therefore, we only focus on the conditional mutual information, where
\begin{equation}
\begin{aligned}
I(Q_{j^i}^{\text{S},+};Q_{j^b}^{\text{S},+}|C) & \\
= \sum_{c_j, q_{i,j}^{\text{S},+},q_{b,j}^{\text{S},+}} &P(c_{j}, q_{i,j}^{\text{S},+}, q_{b,j}^{\text{S},+}) \\
&\log \frac{P(q_{b,j}^{\text{S},+}q_{i,j}^{\text{S},+}|c_{j})}{P(q_{b,j}^{\text{S},+}|c_{j})P(q_{i,j}^{\text{S},+}|c_{j})}.
\label{eq:MI_cate_condition}
\end{aligned}
\end{equation}
We decompose Eq. \ref{eq:MI_cate_condition} to obtain:
\begin{equation}
\begin{aligned}
&I(Q_{j^i}^{\text{S},+};Q_{j^b}^{\text{S},+}|C) \\
&= \sum_{c_j, q_{i,j}^{\text{S},+},q_{b,j}^{\text{S},+}} P(c_{j}, q_{i,j}^{\text{S},+}, q_{b,j}^{\text{S},+}) \log P(q_{b,j}^{\text{S},+}q_{i,j}^{\text{S},+}|c_{j}), \\
&- \sum_{c_j, q_{i,j}^{\text{S},+},q_{b,j}^{\text{S},+}} P(c_{j}, q_{i,j}^{\text{S},+}, q_{b,j}^{\text{S},+}) \log P(q_{b,j}^{\text{S},+}|c_{j}), \\
&- \sum_{c_j, q_{i,j}^{\text{S},+},q_{b,j}^{\text{S},+}} P(c_{j}, q_{i,j}^{\text{S},+}, q_{b,j}^{\text{S},+}) \log P(q_{i,j}^{\text{S},+}|c_{j}).
\label{eq:MI_cate_condition_decompose}
\end{aligned}
\end{equation}
Additionally, $I(C; Q_{j^i}^{\text{S},+})$ also can be written as:
\begin{equation}
\begin{aligned}
&I(C; Q_{j^i}^{\text{S},+}) =\sum_{q_{i,j}^{\text{S},+}} \sum_{c_j} P(q_{i,j}^{\text{S},+}, c_{j}) \log \frac{P(q_{i,j}^{\text{S},+}|c_j)}{P(q_{i,j}^{\text{S},+})}, \\
&= \sum_{c_j, q_{i,j}^{\text{S},+},q_{b,j}^{\text{S},+}} P(c_{j}, q_{i,j}^{\text{S},+}, q_{b,j}^{\text{S},+}) \log \frac{P(q_{i,j}^{\text{S},+}|c_j)}{P(q_{i,j}^{\text{S},+})}.
\label{eq:MI_cate_Qi_C}
\end{aligned}
\end{equation}
Similarly, we decompose Eq. \ref{eq:MI_cate_Qi_C} to obtain:
\begin{equation}
\begin{aligned}
&I(C; Q_{j^i}^{\text{S},+}) \\
&= \sum_{c_j, q_{i,j}^{\text{S},+},q_{b,j}^{\text{S},+}} P(c_{j}, q_{i,j}^{\text{S},+}, q_{b,j}^{\text{S},+}) \log P(q_{i,j}^{\text{S},+}|c_j), \\
&- \sum_{c_j, q_{i,j}^{\text{S},+},q_{b,j}^{\text{S},+}} P(c_{j}, q_{i,j}^{\text{S},+}, q_{b,j}^{\text{S},+}) \log {P(q_{i,j}^{\text{S},+})}. \\
\label{eq:MI_cate_Qi_C_decompose}
\end{aligned}
\end{equation}
Since the equation:
\begin{align}
I(Q_{j^i}^{\text{S},+};Q_{j^b}^{\text{S},+}|C) = I(C, Q_{j^b}^{\text{S},+}; Q_{j^i}^{\text{S},+}) - I(C; Q_{j^i}^{\text{S},+})
\end{align}
holds, substituting Eq. \ref{eq:MI_cate_decompose}, Eq. \ref{eq:MI_cate_condition_decompose}, and Eq. \ref{eq:MI_cate_Qi_C_decompose} into it yields:
\begin{equation}
\begin{aligned}
&\sum_{c_j, q_{i,j}^{\text{S},+},q_{b,j}^{\text{S},+}} \log P(c_{j}, q_{i,j}^{\text{S},+}, q_{b,j}^{\text{S},+})P(q_{b,j}^{\text{S},+}q_{i,j}^{\text{S},+}|c_{j}) \\
&= \sum_{c_j, q_{i,j}^{\text{S},+},q_{b,j}^{\text{S},+}} P(c_{j}, q_{i,j}^{\text{S},+}, q_{b,j}^{\text{S},+}) \log P(q_{i,j}^{\text{S},+}|c_{j} q_{b,j}^{\text{S},+}), \\
&+ \sum_{c_j, q_{i,j}^
{\text{S},+},q_{b,j}^{\text{S},+}} P(c_{j}, q_{i,j}^{\text{S},+}, q_{b,j}^{\text{S},+}) \log P(q_{b,j}^{\text{S},+}|c_{j}).
\label{eq:eqention}
\end{aligned}
\end{equation}
Then, we substitute  Eq. \ref{eq:eqention} into Eq. \ref{eq:MI_cate_condition_decompose} and apply Jensen's inequality, resulting in: 
\begin{equation}
\begin{aligned}
&I(Q_{j^i}^{\text{S},+};Q_{j^b}^{\text{S},+}|C) \\
&= \sum_{c_j, q_{i,j}^{\text{S},+},q_{b,j}^{\text{S},+}} P(c_{j}, q_{i,j}^{\text{S},+}, q_{b,j}^{\text{S},+}) \log P(q_{i,j}^{\text{S},+}|c_{j} q_{b,j}^{\text{S},+}) \\
&- \sum_{c_j, q_{i,j}^{\text{S},+},q_{b,j}^{\text{S},+}} P(c_{j}, q_{i,j}^{\text{S},+}, q_{b,j}^{\text{S},+}) \log P(q_{i,j}^{\text{S},+}|c_{j}), \\
&= \sum_{c_j, q_{i,j}^{\text{S},+},q_{b,j}^{\text{S},+}} P(c_{j}, q_{i,j}^{\text{S},+}, q_{b,j}^{\text{S},+}) \log \frac{P(q_{i,j}^{\text{S},+}|c_{j} q_{b,j}^{\text{S},+})}{P(q_{i,j}^{\text{S},+}|c_{j})}, \\
&\leq \log \sum_{c_j, q_{i,j}^{\text{S},+},q_{b,j}^{\text{S},+}} P(c_{j}, q_{i,j}^{\text{S},+}, q_{b,j}^{\text{S},+}) \frac{P(q_{i,j}^{\text{S},+}|c_{j} q_{b,j}^{\text{S},+})}{P(q_{i,j}^{\text{S},+}|c_{j})}, \\
&= \log \sum_{c_j, q_{i,j}^{\text{S},+},q_{b,j}^{\text{S},+}} P(c_{j}) P(q_{i,j}^{\text{S},+}|c_{j} q_{b,j}^{\text{S},+}) P(q_{b,j}^{\text{S},+}|c_{j} q_{i,j}^{\text{S},+}).
\label{eq:MI_cate_condition_simplify_1}
\end{aligned}
\end{equation}
Given that $q_{i,j}^{\text{S},+}$ and $q_{b,j}^{\text{S},+}$ come from the same range, $P(q_{i,j}^{\text{S},+}|c_{j} q_{b,j}^{\text{S},+}) \propto P(q_{b,j}^{\text{S},+}|c_{j} q_{i,j}^{\text{S},+})$.
By ignoring the fixed distribution $P(c_{j})$, optimizing $I(Q_{j^i}^{\text{S},+};Q_{j^b}^{\text{S},+}|C)$ depends on the conditional probability $P(q_{i,j}^{\text{S},+}|c_{j} q_{b,j}^{\text{S},+})$, which is defined as:
\begin{align}
P(q_{i,j}^{\text{S},+}|c_{j} & q_{b,j}^{\text{S},+})= \\
&\frac{\sum_{b=1}^{B^+}\text{sim}(q_{i,j}^{\text{S},+}, q_{b,j}^{\text{S},+})}{\sum_{b=1}^{B}\text{sim}(q_{i,j}^{\text{S},+}, q_{b,j}^S) + \text{sim}(q_{i,j}^{\text{S},-}, c_j)}.
\end{align}
 For the convenience of expression, we split the denominator into positive and negative terms according to the term of numerator, yielding:
\begin{align}
&\Delta({pos}) = \sum\nolimits_{b=1}^{B^+}\text{sim}(q_{i,j}^{\text{S},+}, q_{b,j}^{\text{S},+}), \label{eq:pos_cate2}\\
&\Delta(neg) = \sum\nolimits_{b=1}^{B^-}\text{sim}(q_{i,j}^{\text{S},+}, q_{b,j}^{\text{S},-}) + \text{sim}(q_{i,j}^{\text{S},-}, c_j).
\label{eq:neg_cate2}
\end{align}
Here, $\text{sim}(.,.)$ in $\mathcal{L}_{\text{cst}}^\text{cate.}$ is substituted to $f(.,.)$. 

\section{C.Fine-tuning Datasets.}
We evaluate our model using three CXR datasets:
(i) \textbf{\textit{Child}CXRs}, which contains healthy and pneumonia-related anomaly images. This dataset is officially divided into training (1,349 normal images and 3,883 disease images) and testing subsets (234 normal images and 390 disease images).
(ii) \textbf{NIH}, which includes 112,120 frontal-view CXR images with 14 disease labels obtained from 30,805 unique patients, where 14 disease labels including Atelectasis (Atel.), Cardiomegaly (Card.), Effusion (Effu.), Infiltration (Infi.), Mass, Nodule (Nodu.), Pneumonia (Pneu.), Pneumothorax (PneuX.), Consolidation (Cons.), Edema (Edem.), Emphysema (Emph.), Fibrosis (Fibr.), Pleural Thickening (P.T.), and Hernia (Hern.). For multi-label classification experiments, we use the both official partitioning and five-fold cross-validation scheme. For anomaly detection, we adhere to the official division but categorize images into normal and abnormal.
(iii) \textbf{CXP}, comprising 224,316 CXR images from 65,240 patients. We utilize the official validation set as the test set. In line with the original study, we perform multi-label classification for Atelectasis (Atel.), Cardiomegaly (Card.), Edema, Consolidation (Cons.), and Pleural Effusion (P.E.) with both official partitioning and five-fold cross-validation scheme. For anomaly detection, following the SQUID, we use 5,249 normal for training, and a random selection of 250 normal and 250 abnormal images for testing.


\begin{table}[t]
  \centering
  \resizebox{\columnwidth}{!}{%
    \begin{tabular}{cccc|ccc}
    \toprule
    \# & C.P.        & $z_{mask}$     & $I(Q_{b,j}^{\text{S},-}, Q_{i,j}^{\text{S},-})$ & AUC & ACC & F1 \\ \midrule
    1 & Probability  & \CheckmarkBold & \XSolidBrush                      & \(\mathbf{89.4}\) & \(\mathbf{83.8}\) & \(\mathbf{84.5}\) \\
    2 & Feature      & \CheckmarkBold & \XSolidBrush                      & 84.8 & 81.6 & 80.9 \\
    3 & Probability  & \XSolidBrush   & \XSolidBrush                      & 85.7 & 80.4 & 81.2 \\
    4 & Probability  & \CheckmarkBold & \CheckmarkBold                    & 77.9 & 70.4 & 72.4 \\ 
    \bottomrule
    \end{tabular}%
  }
  \caption{\textbf{Impact of different components.} AUC scores $\uparrow$ (\%), ACC scores $\uparrow$ (\%), F1 scores $\uparrow$ (\%) on \textit{Child}CXR. }
  \label{tab:Ablation_expand}
\end{table}

\begin{table}[t]
\centering
\resizebox{\columnwidth}{!}{%
\begin{tabular}{c|cccccccccccccc|c}
\toprule
   & Atel.& Card.& Cons.& Edema& E.C. & Frac.& LungL.& LungO.& N.F.& P.E.& P.O.& Pneu. & PneuX.& S.D.& Number \\\midrule
1  & 0    & 0    & 0    & 0    & 0    & 0    & 0    & 0    & 1  & 0     & 0     & 0    & 0     & 0   & 69,540 \\
2  & 0    & 0    & 0    & 0    & 0    & 0    & 0    & 0    & 1  & 0     & 0     & 0    & 0     & 1   & 11,577  \\
3  & 0    & 0    & 0    & 0    & 0    & 0    & 0    & 0    & 0  & 0     & 0     & 0    & 0     & 0   & 11,303  \\
4  & 0    & 0    & 0    & 0    & 0    & 0    & 0    & 1    & 0  & 0     & 0     & 0    & 0     & 0   & 10,123  \\
5  & 0    & 1    & 0    & 0    & 0    & 0    & 0    & 0    & 0  & 0     & 0     & 0    & 0     & 0   & 7,679   \\
6  & 1    & 0    & 0    & 0    & 0    & 0    & 0    & 0    & 0  & 0     & 0     & 0    & 0     & 0   & 5,525   \\
7  & 0    & 0    & 0    & 0    & 0    & 0    & 0    & 0    & 0  & 1     & 0     & 0    & 0     & 0   & 4,922   \\
8  & 0    & 1    & 0    & 0    & 0    & 0    & 0    & 0    & 0  & 0     & 0     & 0    & 0     & 1   & 4,170   \\
9  & 0    & 0    & 0    & 0    & 0    & 0    & 0    & 1    & 0  & 0     & 0     & 0    & 0     & 1   & 3,648   \\
10 & 1    & 0    & 0    & 0    & 0    & 0    & 0    & 0    & 0  & 1     & 0     & 0    & 0     & 0   & 3,507   \\
11 & 1    & 0    & 0    & 0    & 0    & 0    & 0    & 0    & 0  & 0     & 0     & 0    & 0     & 1   & 3,205   \\
12 & 0    & 0    & 0    & 1    & 0    & 0    & 0    & 0    & 0  & 0     & 0     & 0    & 0     & 0   & 3,155   \\
13 & 0    & 0    & 0    & 0    & 0    & 0    & 0    & 0    & 0  & 0     & 0     & 1    & 0     & 0   & 3,064   \\
14 & 0    & 0    & 0    & 0    & 0    & 0    & 0    & 0    & 0  & 0     & 0     & 0    & 0     & 1   & 3,022   \\
15 & 1    & 0    & 0    & 0    & 0    & 0    & 0    & 0    & 0  & 1     & 0     & 0    & 0     & 1   & 2,742   \\
16 & 1    & 0    & 0    & 0    & 0    & 0    & 0    & 1    & 0  & 0     & 0     & 0    & 0     & 0   & 2,594   \\
17 & 0    & 0    & 0    & 0    & 0    & 0    & 0    & 0    & 0  & 1     & 0     & 0    & 0     & 1   & 2,233   \\
18 & 0    & 0    & 0    & 0    & 0    & 0    & 0    & 1    & 0  & 0     & 0     & 1    & 0     & 0   & 2,191   \\
19 & 1    & 1    & 0    & 0    & 0    & 0    & 0    & 0    & 0  & 1     & 0     & 0    & 0     & 1   & 2,163   \\
20 & 0    & 0    & 0    & 0    & 0    & 0    & 0    & 1    & 0  & 1     & 0     & 0    & 0     & 0   & 2,079   \\
\bottomrule
\end{tabular}%
}
\caption{top 20 categories with the highest occurrences in MIMIC-CXR-JPG dataset.}
\label{tab:mimic_count}
\end{table}

\section{D.Training Settings.}
For model evaluation, we follow the \textit{pre-training (on source data)} $\rightarrow$ \textit{fine-tuning (on target data)} protocol and employ two settings: transfer learning and anomaly detection. In these settings, the source and target data are derived from different datasets. Specifically, in the first setting, we fine-tune the pre-trained encoder of AFiRe with both limited and full annotations, applying data augmentations such as random resize crops and horizontal flips. The performance is evaluated by reporting the AUC score based on the global average pool of the extracted image token sequence.

For the evaluations of anomaly detection, we initialize the whole network with the pre-trained weights and adapt it to the target datasets using the proposed SSL task. 
During the testing phase, we define the anomaly score as follows: 
\begin{align}
\mathcal{A}(x_i) = (\frac{1}{L}\sum_{j=1}^L \exp (\|y_{i,j}-x_{i,j}\|^2))^{-1},
\label{eq: anomaly}
\end{align}
where $\|y_{i,j}-x_{i,j}\|^2$ denotes the squared Euclidean distances between the restored normal image patch $y_{i,j}$ and the input patch $x_{i,j}$. $L$ is the length of the image patches.
Due to the characteristics of the exponential function, when $x_{i,j}$ and $y_{i,j}$ are very similar, the value of $|y_{i,j}-x_{i,j}|^2$ approaches zero, resulting in $\exp(0)=1$. Conversely, when $x_{i,j}$ and $y_{i,j}$ are significantly different, the values of $|y_{i,j}-x_{i,j}|^2$ become large, leading to $\exp(\infty)=\infty$. Therefore, we take the reciprocal of the squared Euclidean distances to ensure the anomaly value is larger for more similar images and smaller for less similar images.

\textbf{Pre-training settings:} 
Unlike other global-wise instance contrastive learning models, in the pre-training stage, we do not apply conventional image transformations, such as rotation, affine, and random crop, which can corrupt anatomic structure. Instead, for AFiRe, we utilize the proposed Synthetic Lesion Mask as a form of data augmentation. These images are normalized using the mean and standard deviation obtained from the ImageNet dataset.

We use the Vision Transformer (ViT) as the backbone for our proposed self-supervised tasks. Specifically, we employ ViT-B as the encoder, a lightweight ViT architecture, which with 8 layers and 512 embedding dimensions, as the decoder. The projection head, $h_\theta$ consists of a $MLP$ layer with 256 hidden dimensions. The EMA factor $\lambda$ for updating the momentum encoder is set to 0.99 and is increased to 1 according to a cosine schedule during training.

In the pre-training phase, we use the AdamW optimizer with a weight decay of 0.04. Transformer blocks are initialized using Xavier uniform initialization. The learning rate ($lr$) is linearly ramped up during the first 20 epochs to a base value determined by the linear scaling rule: $lr = 0.0005 \times \frac{\text{batch-size}}{256}$. After this warm-up period, we apply a cosine annealing strategy to decrease the learning rate
Weight decay is adjusted according to a cosine schedule, ranging from 0.04 to 0.4. The temperature $\tau_s$ is set to 0.1, while $\tau_t$ undergoes a linear warm-up from 0.04 to 0.07 over the first 30 epochs.
For other hyperparameters, we empirically determine the number of synthetic lesions in $M_i^n$ to be $R=[1,4]$, based on the statistics in Table \ref{tab:mimic_count}, which indicate that lesions generally appear in up to 4 types. The number of $M_i^n$ for a normal radiographic image is set to $N=9$ to balance the availability of pre-training data with the duration of training. We implement our pre-training model using PyTorch and distribute the training across 8 NVIDIA A6000 GPUs, running for 800 epochs with a batch size of 64 for 6 days.

\begin{figure*}[h]
  \centering
  \includegraphics[width=1\linewidth]{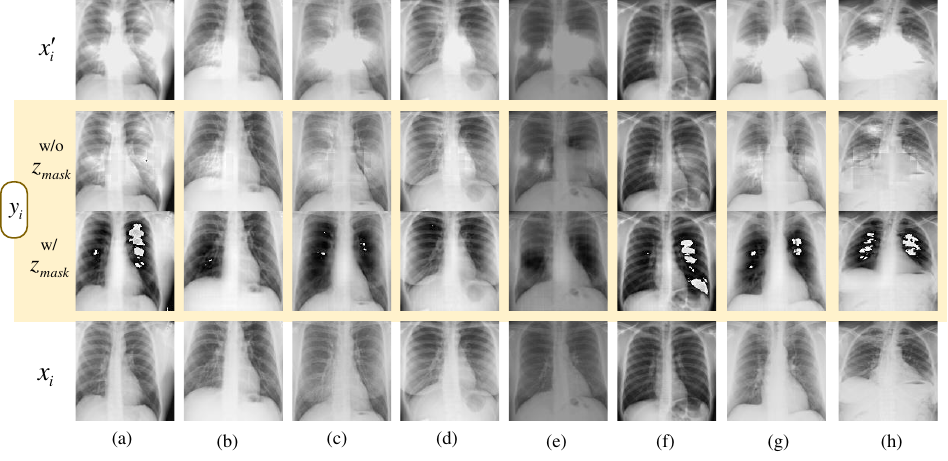}
  \caption{Visualiztion of $y_i$ with or without using $z_{mask}$ when the input is synthetic abnormal image ($x^{\prime}_i$).} 
  \label{fig:generate_abnormal}
\end{figure*}

\begin{figure*}[ht]
  \centering
  \includegraphics[width=1\linewidth]{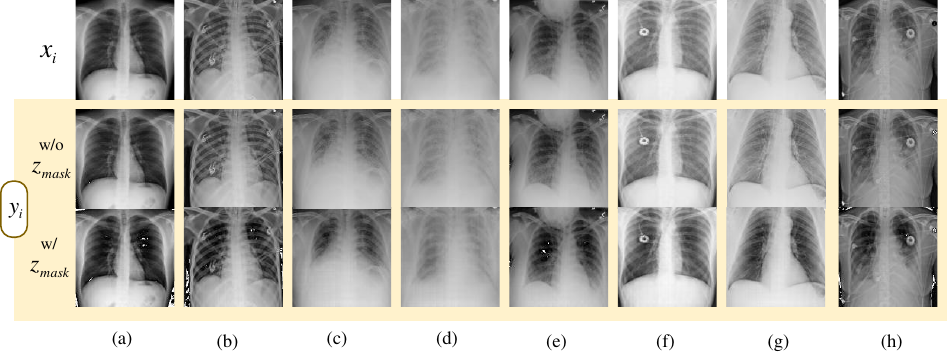}
  \caption{Visualiztion of $y_i$ with or without using $z_{mask}$ when the input is real normal image ($x_i$).} 
  \label{fig:generate_normal}
\end{figure*}

\textbf{Fine-tuning settings:}
Following the standard transfer learning protocol, we fine-tune the pre-trained encoder $\mathbf{E}_\text{T}$ for multi-label classification by adding a task-specific head. Additionally, we fine-tune the pre-trained encoders $\mathbf{E}_\text{S}$ and $\mathbf{E}_\text{T}$, as well as the decoder $\mathbf{R}$, for anomaly detection tasks.
To evaluate the generalization of AFiRe representations, we fine-tune all parameters of the downstream models and assess their performance using the following metrics:
(i) Area Under the ROC Curve (AUC) for multi-label classification.
(ii) AUC, Accuracy (ACC), and F1-score (F1) for anomaly detection.
We optimize each downstream task using the best-performing hyperparameters. For multi-label classification, we use the AdamW optimizer with a learning rate ($lr$) of $2 \times 10^{-4}$ on the NIH dataset and $1 \times 10^{-4}$ on the CXP dataset. Learning rates are decayed using a cosine schedule with a weight decay of 0.05. Early stopping is employed if the metrics do not improve for 10 epochs on the validation set.
For data augmentation, we apply standard techniques including random cropping, flipping, and rotation. These experiments are implemented in PyTorch and are trained on a single NVIDIA A6000 GPU, with a batch size of 128, over a period of 7 hours.
In anomaly detection tasks, we use the Adam optimizer with a learning rate of $1 \times 10^{-4}$ for the \textit{Child}CXRs, NIH, and CXP datasets. The learning rate is decayed using a cosine schedule. Experiments are also implemented in PyTorch and trained on a single NVIDIA A6000 GPU, with a batch size of 64, for total of 12 hours. Other hyperparameters remain the same as those used during the pre-training stage.

For each experiment, we repeat it three times and report their average results. 
We also provide statistical analysis using an independent two-sample t-test in Appendix G and H.

\section{E. Impact of Other Components.}
Table \ref{tab:Ablation_expand} elucidates the impact of additional different components of AFiRe. We assess the linear probing performance on the \textit{Child}CXR dataset, presenting AUC, ACC, and F1 scores. The components under investigation include contrastive protocol in Task \uppercase\expandafter{\romannumeral1} (Probability Contrast vs. Feature Contrast), the integration of $z_{mask}$, and the additional mutual information between abnormal tokens ($I(Q_{j^b}^{\text{S},-}, Q_{j^i}^{\text{S},-})$). Row 1 demonstrates the default setting in AFiRe. 

\textit{\textbf{Alignment of distinct token features in unique probability space effectively mitigates discreteness arising from individual differences.}}
In different contrastive protocol choices (row 2), we observe that classification performance presents a decrease when directly contrasting the token features($z_{i,j}$) rather than token probabilities ($q_{i,j}$). This result reveals that mapping features to a unified probability distribution space effectively reduces the discreteness caused by individual differences. In this unified space, the similarity between different features becomes more intuitive, as their higher-level contextual semantics can be explicitly represented through independent probabilistic attributes. This probabilistic representation offers richer discriminative information compared to individual feature values, helping to prevent the model from prematurely converging to a local optimum during training. Consequently, leveraging token probabilities to maximize mutual information can improve both model performance and stability.


\textit{\textbf{Substituting abnormal token features with $z_{mask}$ effectively preserves both normal and abnormal information while restoring fine-grained details closely associated with various anatomical structures.}} In row 3 of Table \ref{tab:Ablation_expand}, we present the quantification results when $z_{mask}$ is absent, \textit{i.e.}, when anomalies were not removed during the restoration of latent features to their normal anatomical shape. To illustrate the impact of $z_{mask}$ more intuitively in the reconstruction task, we visually compared the generated $y_i$ with and without $z_{mask}$ using synthetic abnormal radiography images ($x^{\prime}_i$) in Fig. \ref{fig:generate_abnormal} and normal images ($x_i$) in Fig. \ref{fig:generate_normal}. Our analysis, considering both classification and restoration outcomes, indicates that replacing abnormal tokens with $z_{mask}$ not only enhances quantification performance but also improves visualization results to remove anomalies and restore them to their normal shape. Specifically, using $z_{mask}$ improves classification performance, increasing ACC from 85.7\% to 89.4\%, AUC from 80.4\% to 83.8\%, and F1 from 81.2\% to 84.5\%. The more pronounced results are illustrated in Fig. \ref{fig:generate_abnormal} and Fig. \ref{fig:generate_normal}. 
For example, when the input image is $x^{\prime}_i$ (Fig. \ref{fig:generate_abnormal}), the model with $z_{mask}$ often performs better in both removing anomalies and restoring these local areas to their normal shape, as shown in rows (b) and (d). In contrast, directly restoring the latent token features with the decoder (without $z_{mask}$), when anomalies are significantly prominent (generally do not match the anatomical patterns of lesions in reality), can reduce the impact of abnormal shadows to some extent, as seen in rows (a), (c), (e), (g), and (h). However, it still cannot effectively identify the correct anatomical structure in abnormal regions compared to using $z_{mask}$.
Additionally, we observe a common phenomenon with $z_{mask}$: the generated images often show damage in areas where the density of the anomalous region covers the original anatomical structure. This phenomenon also extends to normal input images (Fig. \ref{fig:generate_normal}). In other words, the generated normal image tends to have a lower density in the chest region compared to the original image. However, the performance with $z_{mask}$ is generally better than without $z_{mask}$.

\textit{\textbf{Enforcing consistency among negative tokens limits the model's ability to generalize and recognize abnormal semantic information.}} In the context of radiography images, pathologies exhibit significant diversity and irregularity. Pathological variations are often subtle and can manifest in numerous ways across different cases. For example, a tumor might appear as a slight shadow in one image and a distinct mass in another, depending on factors like size, location, and the patient's anatomy. In row 4 of Table \ref{tab:Ablation_expand}, we maintain consistency among abnormal tokens in AFiRe, which leads to reduced accuracy in differentiating between various pathological conditions. This ultimately indicates that enforcing consistency among negative tokens could limit the model's ability to generalize and recognize abnormal semantic information.

\begin{table*}[t]
\centering
\resizebox{1\textwidth}{!}{%
\begin{tabular}{c|l|ccccccccccccccc|cccccc}
\toprule
\multirow{2}{*}{ratio}   & \multirow{2}{*}{Model} & \multicolumn{15}{c|}{NIN}                                                                                                                                                                                     & \multicolumn{6}{c}{CXP} \\\cline{3-23}

                         &                        &Atel.        &Card.        &Effu.        &Infi.        &Mass        &Nodu.        &Pneu.	      &PneuX.       &Cons.        &Edem.	    &Emph.	      &Fibr.        &P.T.       & \multicolumn{1}{c|}{Hern.} & Mean   & Atel.        &Card.         &Edem.        &Cons.	      &\multicolumn{1}{c|}{P.E.}              &Mean\\ \midrule
                   
\multirow{9}{*}{$1\%$}   &{ViTB \cite{vit}}       &56.9         &60.2         &62.3         &58.3         &54.5         &52.6         &54.4         &54.4         &56.9         &61.5         &51.6         &64.5         &58.8         & \multicolumn{1}{c|}{52.3}  &57.1 & 73.3         &67.6          &85.3         &64.4         &\multicolumn{1}{c|}{84.9}              &75.1      \\
                         &{MoCov2 \cite{mocoV2}}  &58.5         &63.8         &65.5         &60.6         &56.2         &54.5         &55.5         &57.5         &57.6         &65.1         &53.5         &66.4         &58.3         & \multicolumn{1}{c|}{55.6}  &59.2 & 73.5         &\textbf{70.2} &85.8         &65.6         &\multicolumn{1}{c|}{85.8}              &76.2   \\
                         &{DINO \cite{dino}}      &58.8         &64.2         &66.3         &60.9         &56.0         &55.3         &56.5         &59.6         &\textbf{60.8}&67.4         &54.9         &68.1         &61.7         & \multicolumn{1}{c|}{57.9}  &60.6 & 74.4         &69.8          &86.4         &69.2         &\multicolumn{1}{c|}{84.7}              &76.9      \\
                         &{MAE \cite{MAE}}        &60.5         &64.3         &66.6         &59.3         &55.3         &56.7         &55.7         &60.7         &59.4         &67.6         &54.2         &68.3         &62.0         & \multicolumn{1}{c|}{56.4}  &60.5 & \textbf{74.8}&69.8          &\textbf{86.8}&70.5         &\multicolumn{1}{c|}{86.6}              &77.7     \\
                         &{C2L \cite{C2L}}        &61.1         &66.5         &67.5         &63.1         &56.1         &56.0         &57.5         &61.0         &59.1         &68.4         &56.1         &\textbf{68.8}&65.4         & \multicolumn{1}{c|}{57.2}  &61.7 & 73.9         &68.3          &82.9         &76.4         &\multicolumn{1}{c|}{87.1}              &77.6      \\
                         &{TransVW \cite{TransVW}}&60.8         &65.4         &67.0         &62.6         &55.7         &56.9         &57.8         &60.9         &58.4         &66.9         &55.3         &67.9         &65.3         & \multicolumn{1}{c|}{55.9}  &61.2 & \textbf{74.8}&69.2          &83.6         &76.6         &\multicolumn{1}{c|}{86.8}              &78.2    \\
                         &{DiRA \cite{Dirav2}}      &61.5         &67.8         &69.6         &63.4         &58.3         &\textbf{57.2}&58.8         &61.3         &60.4         &69.7         &56.5         &68.4         &\textbf{66.7}& \multicolumn{1}{c|}{57.3}  &62.6  & 74.4         &68.7          &84.1         &77.4        &\multicolumn{1}{c|}{87.3}              &78.4   \\
                         &{PCRL \cite{PCRL}}      &\textbf{62.7}&68.3         &\textbf{70.4}&62.8         &\textbf{60.5}&55.2         &59.2         &\textbf{61.6}&60.1         &69.3         &57.3         &68.0         &66.2         & \multicolumn{1}{c|}{\textbf{59.1}} &62.9  &74.6	&68.5	&83.7	&77.1	&\multicolumn{1}{c|}{86.8}  &78.1   \\
                         &{AFiRe(ours)}            &62.2         &\textbf{69.4}&68.7         &\textbf{63.8}&60.4         &56.1         &\textbf{59.9}&61.4         &60.4         &\textbf{70.5}&\textbf{58.0}&68.4         &66.0         & \multicolumn{1}{c|}{\textbf{59.1}}&\textbf{63.2} & 74.4         &69.5          &84.5         &\textbf{78.2}&\multicolumn{1}{c|}{\textbf{87.4}}     &\textbf{78.8}\\ \hline
                         
\multirow{9}{*}{$10\%$}  &{ViTB \cite{vit}}       &66.4         &59.3         &72.9         &56.6         &71.3         &59.5         &65.7         &74.9         &71.8         &79.4         &76.1         &62.8         &67.1         & \multicolumn{1}{c|}{61.4}  &67.5  & 80.2         &82.5          &89.6         &78.6         &\multicolumn{1}{c|}{89.2}              &84.0   \\
                         &{MoCov2 \cite{mocoV2}}  &67.4         &61.4         &74.9         &59.8         &72.1         &63.4         &67.9         &76.3         &72.4         &79.8         &75.9         &63.9         &68.2         & \multicolumn{1}{c|}{61.9}  &68.9  & 79.9         &82.2          &88.8         &81.9         &\multicolumn{1}{c|}{88.5}              &84.3   \\
                         &{DINO \cite{dino}}      &68.9         &62.3         &76.5         &59.7         &72.9         &66.7         &67.4         &76.8         &71.3         &80.4         &74.8         &64.7         &69.8         & \multicolumn{1}{c|}{62.9}  &69.6  & 80.4         &82.6          &90.0         &85.2         &\multicolumn{1}{c|}{86.4}              &84.9   \\
                         &{MAE \cite{MAE}}        &68.4         &63.7         &76.9         &60.4         &73.2         &65.7         &67.9         &76.2         &72.6         &81.9         &76.2         &63.9         &70.5         & \multicolumn{1}{c|}{63.8}  &70.1  & 80.6         &\textbf{83.4}&\textbf{91.2}&84.8         &\multicolumn{1}{c|}{86.1}              &85.2   \\
                         &{C2L \cite{C2L}}        &72.5         &68.0         &81.3         &62.4         &75.8         &67.2         &70.2         &80.6         &74.8         &85.4         &78.4         &68.3         &72.2         & \multicolumn{1}{c|}{66.1}  &73.1  & 80.4         &80.3          &87.4         &88.6         &\multicolumn{1}{c|}{90.3}              &85.4    \\
                         &{TransVW \cite{TransVW}}&69.4         &64.3         &78.2         &59.5         &72.6         &63.1         &67.2         &77.2         &70.9         &83.0         &75.3         &65.8         &68.9         & \multicolumn{1}{c|}{60.2}  &69.7  & 79.8         &78.6          &88.2         &87.2         &\multicolumn{1}{c|}{88.6}              &84.5   \\
                         &{DiRA \cite{Dirav2}}      &72.4         &68.7         &83.2         &64.7         &78.3         &67.5         &70.6         &84.3         &77.2         &87.7         &79.5         &70.8         &75.2         & \multicolumn{1}{c|}{68.1}  &74.9  & 80.2         &79.4          &86.5         &89.4         &\multicolumn{1}{c|}{\textbf{90.6}}     &85.2   \\
                         &{PCRL \cite{PCRL}}      &\textbf{75.4}&70.6         &\textbf{84.2}&\textbf{65.5}&78.9         &69.6         &72.7         &83.5         &\textbf{77.6}&88.5         &\textbf{80.8}&71.3         &74.8         & \multicolumn{1}{c|}{67.6}  & 75.8 &81.1	&79.6	&87.5	&90.3 	&\multicolumn{1}{c|}{88.9}  &85.5    \\
                         &{AFiRe(ours)}            &75.0         &\textbf{71.4}&82.9         &\textbf{65.5}&\textbf{79.0}&\textbf{70.5}&\textbf{73.2}&\textbf{84.9}&76.5         &\textbf{89.4}&80.2         &\textbf{71.9}&\textbf{75.6}& \multicolumn{1}{c|}{\textbf{69.6}}  &\textbf{76.1} & \textbf{81.6}&80.3          &87.9         &\textbf{91.2}&\multicolumn{1}{c|}{\textbf{90.6}}     &\textbf{86.3}\\ \hline
                         
\multirow{8}{*}{$100\%$}&{ViTB \cite{vit}}        &79.1         &72.5         &86.7         &68.7         &83.5         &72.7         &79.0         &86.6         &\textbf{85.0}&91.2         &87.7         &75.3         &80.1         & \multicolumn{1}{c|}{73.8}  &80.1 & 83.2         &\textbf{86.3}&93.4         &82.8         &\multicolumn{1}{c|}{92.4}              &87.6    \\
                         &{MoCov2 \cite{mocoV2}}  &77.9         &72.6         &84.7         &70.1         &84.8         &75.9         &\textbf{80.6}&\textbf{88.5}&83.1         &91.2         &\textbf{89.0}&75.1         &\textbf{83.6}& \multicolumn{1}{c|}{73.8}  &80.8 & 81.7         &85.4          &92.9         &85.1         &\multicolumn{1}{c|}{90.1}              &87.0    \\
                         &{DINO \cite{dino}}      &79.2         &73.6         &87.4         &70.7         &84.8         &\textbf{78.0}&79.9         &87.6         &81.9         &\textbf{92.0}&86.4         &76.0         &80.5         & \multicolumn{1}{c|}{75.1}  &81.0 & 82.3         &85.4          &93.6         &85.9         &\multicolumn{1}{c|}{90.8}              &87.6    \\
                         &{MAE \cite{MAE}}        &79.3         &74.5         &87.1         &\textbf{71.2}&83.9         &76.7         &78.7         &87.1         &83.6         &90.2         &87.3         &77.4         &81.1         & \multicolumn{1}{c|}{74.3}  &80.9 & 81.9         &86.2          &\textbf{94.9}&88.4         &\multicolumn{1}{c|}{88.7}              &88.0    \\
                         &{C2L \cite{C2L}}        &78.2         &87.0         &85.3         &68.3         &84.8         &73.7         &74.8         &85.5         &79.6         &90.1         &86.3         &80.0         &78.6         & \multicolumn{1}{c|}{88.1}  &81.4 & 83.3         &83.0          &92.7         &93.6         &\multicolumn{1}{c|}{93.8}              &89.3     \\
                         &{MG \cite{MG}}          &77.8         &86.3         &84.7         &67.3         &83.6         &73.0         &74.1         &84.9         &78.8         &89.5         &85.7         &79.6         &78.2         & \multicolumn{1}{c|}{87.0}  &80.8 & 80.6         &81.0          &91.9         &92.7         &\multicolumn{1}{c|}{91.1}              &87.5    \\
                         &{TransVW \cite{TransVW}}&77.9         &86.4         &85.3         &67.6         &84.3         &73.8         &74.4         &85.1         &79.3         &89.8         &86.2         &80.0         &78.6         & \multicolumn{1}{c|}{88.8}  &81.2 & 82.4         &81.8          &93.1         &92.7         &\multicolumn{1}{c|}{91.2}              &88.2   \\
                         &{DiRA \cite{Dirav2}}      &79.6         &87.8         &86.6         &69.4         &85.7         &74.2         &77.8         &86.3         &81.6         &90.3         &88.5         &80.4         &79.7         & \multicolumn{1}{c|}{89.4}  &82.7 & 82.4         &82.7          &92.1         &94.1         &\multicolumn{1}{c|}{\textbf{94.3}}     &87.6   \\
                         &{PCRL \cite{PCRL}}      &\textbf{79.8}&\textbf{88.5}&87.1         &69.7         &86.1         &75.6         &76.1         &87.0         &81.2         &91.6         &87.7         &\textbf{81.7}&80.4         & \multicolumn{1}{c|}{90.2}  &83.0 & 80.5	&82.3	&93.7	&90.4	&\multicolumn{1}{c|}{92.7}  &87.9    \\
                         &{Adamv1 \cite{Adamv1}}    &77.1 	&85.6 	&84.9 	&68.6 	&83.3 	&71.7 	&73.6 	&85.3 	&78.1 	&89.8 	&87.5 	&80.9 	&80.1 	 &\multicolumn{1}{c|}{90.0} &81.2 & 81.3         &80.8          &90.4         &93.1         &\multicolumn{1}{c|}{92.7}     &87.7 \\
                         &{Adamv2 \cite{Adamv2}}    &80.4 	&87.5 	&86.0 	&70.5 	&84.9 	&74.6 	&75.2 	&85.8 	&79.0 	&91.5 	&87.4 	&79.0 	&78.3 	&\multicolumn{1}{c|}{88.5} &82.1 & 82.6         &81.3          &91.6         &93.7         &\multicolumn{1}{c|}{93.9}     &88.6 \\
                         &{AFiRe(ours)}            &79.1         &87.5         &\textbf{87.9}&69.2         &\textbf{86.9}&75.9         &77.2         &87.2         &82.7         &91.0         &88.3         &80.2         &81.2         & \multicolumn{1}{c|}{\textbf{90.8}}  &\textbf{83.2} & \textbf{83.4}&85.2          &92.8         &\textbf{94.3}&\multicolumn{1}{c|}{92.4}              &\textbf{89.6} \\                         
\bottomrule
\end{tabular}%
}
\caption{\textbf{Transfer learning on NIH dataset and CXP dataset with different labeling ratios.} AUC scores $\uparrow$ (\%) of the multi-label classification are reported. The best results are bolded.}
\label{tab:Transfer_expand}
\end{table*}

\begin{table*}[ht]
  \centering
  \resizebox{1\textwidth}{!}{%
    \begin{tabular}{l|cccccccccccccc|c}
    \toprule
Methods     & Atel. & Card. & Effu. & Infi. & Mass & Nodu. & Pneu. & PneuX. & Cons. & Edem. & Emph. & Fibr. & P.T. & Hern. & Mean \\ \midrule
ViTB        & 70.9  & 88.5  & 79.8  & 66.1  & 84.4 & 70.6  & 75.8  & 80.4   & 83.9  & 88.6  & 87.6  & 78.4  & 76.7 & 87.8  & 79.9 \\
DINO        & 74.6  & 87.4  & 83.5  & 65.3  & 82.3 & 70.4  & 73.2  & 83.9   & 83.7  & 89.8  & 87.8  & 79.3  & 76.9 & 86.5  & 80.3 \\
MAE         & 76.4  & 88.9  & 85.7  & 66.7  & 83.5 & 73.5  & 77.5  & 84.8   & 84.2  & 89.1  & 88.2  & 80.7  & 77.1 & 87.9  & 81.7 \\
C2L         & 77.9  & 86.5  & 87.6  & 68.5  & 85.5 & 71.5  & 77.9  & 86.4   & 84.1  & 90.8  & 85.7  & 81.0  & 77.9 & 88.3  & 82.1 \\
DiRA        & 79.4  & 87.1  & 87.3  & 69.9  & 86.6 & 74.8  & 77.8  & 85.6   & \textbf{85.6}  & 89.7  & 86.2  & \textbf{82.9}  & 80.4 & \textbf{89.1}  & 83.0 \\
PCRL        & 79.8  & 86.9  & 88.1  & 70.1  & \textbf{87.1} & 73.9  & \textbf{78.9}  & 87.8   & 84.3  & 91.5  & 88.6  & 81.2  & 79.7 & 88.0  & 83.3 \\
AFiRe(ours) & \textbf{80.8}  & \textbf{89.4}  & \textbf{89.4}  & \textbf{72.1}  & 85.3 & \textbf{76.2}  & 77.4  & \textbf{89.9}   & 84.4  & \textbf{93.4}  & \textbf{89.6}  & 82.5  & \textbf{83.2} & 88.9  & \textbf{84.5} \\
\bottomrule
\end{tabular}%
}
\caption{The average AUC score for the 14 diseases on the NIH dataset using five-fold cross-validation.}
\label{tab:NIH_disease}
\end{table*}

\begin{table}[h]
  \centering
  \resizebox{\columnwidth}{!}{%
    \begin{tabular}{l|ccccc|c} \toprule
Methods     & Atel. & Card. & Edem. & Cons. & P.E. & Mean  \\ \midrule
ViTB        & 78.6  & 81.3  & 88.6  & 89.7  & 88.4 & 85.3 \\
DINO        & 77.4  & 82.2  & 87.9  & 91.1  & 91.9 & 86.1  \\
MAE         & 80.9  & 80.7  & 89.4  & 89.8  & 91.6 & 86.5 \\
C2L         & 79.2  & 81.6  & 91.1  & 89.1  & 89.7 & 86.1 \\
DiRA        & 80.3  & 82.5  & 90.6  & 91.4  & 91.2 & 87.2  \\
Adam1       & 79.6  & 80.2  & 89.8  & 90.7  & 90.8 & 86.2 \\
Adam2       & 80.4  & 81.9  & 91.4  & 92.4  & 91.5 & 87.5 \\
AFiRe(ours) & \textbf{82.2}  & \textbf{83.3}  & \textbf{92.8}  & \textbf{93.1}  & \textbf{93.5} & \textbf{89.0} \\
\bottomrule
\end{tabular}%
}
\caption{The average AUC score for the 5 diseases on the CXP dataset using five-fold cross-validation.}
\label{tab:CXP_disease}
\end{table}

\section{F.Impact of Cluster Number $K$.}
In this experiment, we leverage publicly available pre-trained parameters of MoCo, DINO, and PCL to extract radiography representations. 
Given the multi-disease nature of radiography images, where conditions such as Atelectasis often co-occur with Pleural effusion, we initially binarize the ground truth multi-labels in the MIMIC-CXR-JPG dataset to enhance the distinction between different categories. For example, in our setting, the one-hot labels $[1, 0, 0, 0, 0, 0, 0, 0, 0, 1, 0, 0, 0, 0]$ and $[1, 0, 0, 0, 0, 0, 0, 1, 0, 0, 0, 0, 0, 0]$ represent different categories. Then we count the number of the same category labels across the entire dataset and select the top 20 categories with the highest occurrences (Table \ref{tab:mimic_count}). Here, the 14 disease labels represent Atelectasis, Cardiomegaly, Consolidation, Edema, Enlarged Cardiomediastinum, Fracture, Lung Lesion, Lung Opacity, No Finding, Pleural Effusion, Pleural Other, Pneumonia, Pneumothorax, and Support Devices, respectively. Since the labels of Support Devices (S.D.) are unrelated to diseases, after merging the same category from Table \ref{tab:mimic_count}, we finally obtain 12 distinct categories. We then randomly select 1,000 images to extract their representations and use t-SNE to visualize these representations, employing color differentiation to distinguish between categories. 

In this stricter clustering setting, we observe that the distinct clusters corresponding to different disease categories are more clearly separated, demonstrating the effectiveness of our method in capturing fine-grained anatomical discrimination. This visualization highlights the capability of our approach to accurately group similar cases together, thus aiding in better diagnostic categorization.

Additionally, in AFiRe, we vary $K$ from $64$ to $256$. Notably, AFiRe treats each disjoint image token as an individual sample to predict the distribution probability. Therefore, the final linear layer in $h_{\theta}$ has to handle a dimension of $196 \times K$ vector rather than a single $cls$ token with the dimension of $K$, For example, if $K=65,536$, as adhered to in DINO's setting, $h_{\theta}$ would necessitate a linear layer of dimension $12,845,056$, which demands substantial computational resources. 
To manage this, we need to reduce the values of $k$ for fast and stable training.

Although we reduce the number of $K$, AFiRe still outperforms other contrastive learning models.
While MoCo effectively captures discriminative information, its reliance on a finite dictionary size may limit its ability to capture fine-grained features. The infrequent update of dictionary features may hinder adaptability to new data, potentially compromising clustering performance.
In contrast, DINO enhances clustering capabilities by employing a projection head to map latent features to a higher-dimensional space. The increased dimensionality facilitates more flexible feature adjustments, making it easier to group similar data and separate dissimilar ones.
Furthermore, PCL assigns several prototypes of different granularity to each instance, leading to better cluster results by identifying representative prototypes of the embeddings.
AFiRe achieves superior clustering performance compared to the aforementioned methods by focusing on token-wise representation contrast through the introduction of spatial-aware prototypes.

\begin{table}[h]
  \centering
  \resizebox{\columnwidth}{!}{%
\begin{tabular}{l|ccccc|ccccc} \toprule
\multirow{2}{*}{Methods} & \multicolumn{5}{c|}{NIH}         & \multicolumn{5}{c}{CXP}          \\ \cline{2-11} 
                         & 1    & 2    & 3    & 4    & 5    & 1    & 2    & 3    & 4    & 5    \\ \midrule
ViTB                     & 77.9 & 78.9 & 82.3 & 78.3 & 82.4 & 84.7 & 83.5 & 85.6 & 89.1 & 83.7 \\
DINO                     & 78.1 & 79.8 & 80.8 & 80.1 & 82.8 & 85.2 & 85.7 & 86.4 & 88.6 & 84.6 \\
MAE                      & 79.6 & 81.0 & 83.2 & 81.2 & 83.6 & 85.9 & 84.8 & 87.1 & 89.4 & 85.2 \\
C2L                      & 80.1 & 81.3 & 82.1 & 82.9 & 84.1 & 85.2 & 86.0 & 86.3 & 88.3 & 84.9 \\
DiRA                     & 80.9 & 82.5 & 83.7 & 83.1 & 84.8 & 87.0 & 85.4 & 88.0 & 89.3 & 86.3 \\
PCRL                     & 82.0 & 81.8 & 83.6 & 83.4 & 85.6 &87.4  & 87.1 & 86.7 & 88.4	& 86.8 \\
Adam1                    & 80.3 & 80.1 & 82.4 & 82.7 & 84.4 & 85.9 & 84.6 & 86.9 & 88.8 & 84.9 \\
Adam2                    & \textbf{83.4} & 82.8 & 85.1 & 83.5 & 85.5 & 87.5 & 86.8 & 87.6 & 90.4 & 85.3 \\
AFiRe(ours)              & {82.5} & \textbf{83.5} & \textbf{85.4} & \textbf{84.3} & \textbf{86.6} & \textbf{88.7} & \textbf{88.0}   & \textbf{89.2} & \textbf{91.6} & \textbf{87.4} \\
\bottomrule
\end{tabular}%
}
\caption{The average AUC score in each iteration by five-fold cross-validation.}
\label{tab:5_mean}
\end{table}

\begin{figure*}[h]
  \centering
  \includegraphics[width=1\linewidth]{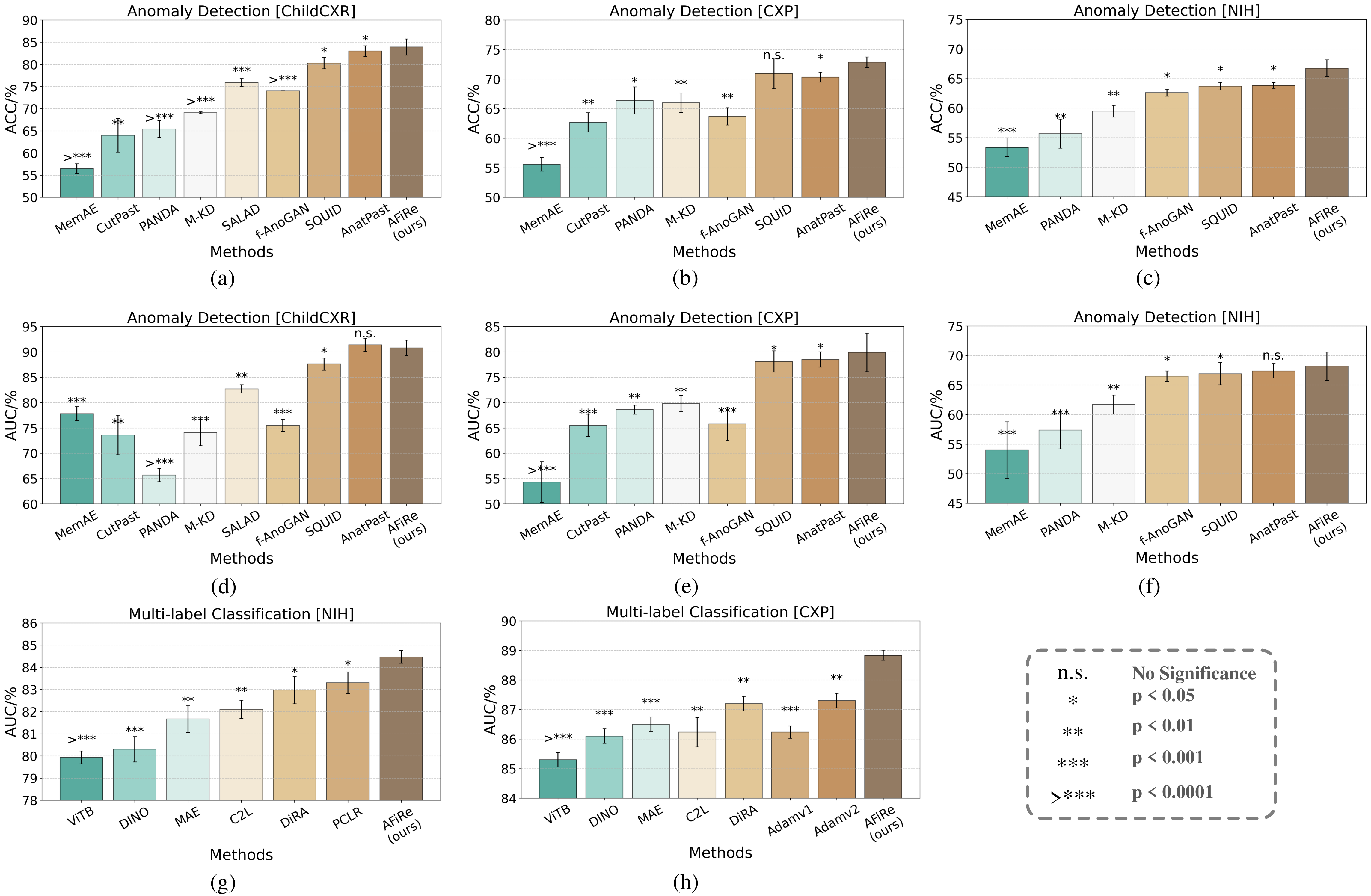}
  \caption{AFiRe provides robust representations, outperforming the state-of-the-art counterparts across different downstream tasks. A statistical significance analysis ($p < 0.05$) was conducted to compare AFiRe with other methods in each task.} 
  \label{fig:p_value}
\end{figure*}

\section{G.More Multi-label Classification Results}
\textit{\textbf{Comparison setting:}} In this experiment, we detail the application of Multi-label Classification on NIH and CXP. Specifically, (i) to assess the generalization performance across various labeling ratios, we use the data split outlined in \cite{PCRL}. (ii) For the five-fold cross-validation, the entire dataset is randomly divided into five equal-sized subsets. In each iteration, AFiRe is trained on four of these subsets and tested on the remaining one. This process is repeated five times, ensuring each subset serves as the test set exactly once. The performance metric (e.g., AUC score) is calculated for each iteration and then averaged to provide a single estimate of the model's performance.

We compare AFiRe with 11 models, including both supervised and self-supervised approaches. These include: (i) 4 baselines (pretrained on ImageNet): ViTB (supervised model),  MoCov2 (feature-contrastive self-supervised model), DINO (probability-contrastive self-supervised model), MAE (restoration-based self-supervised model); (ii) 7 radiography-specific self-supervised methods: C2L, MG, TransVW, DiRA, PCRL, Adamv1, and Adamv2.
Here, C2L is a contrastive-based self-supervised methods. MG and TransVW are restoration-based self-supervised models designed for radiography images. DiRA seamlessly unites discriminative, restorative, and adversarial learning in a unified manner to learn fine-grained semantic representation. PCRL encodes more information into the representations learned from the contrastive loss by reconstructing diverse contexts. Adamv1 and Adamv2 learn anatomy in a coarse-to-fine manner via hierarchical contrastive learning. 
All methods are re-implemented, and experiments are conducted three times to ensure reliability.

\textbf{\textit{Results with limited labeling on NIH Datasets.} }
Left of Table \ref{tab:Transfer_expand} demonstrates the superior efficacy of our proposed model relative to existing state-of-the-art methods. When only $1\%$ of the labeled data is available, AFiRe outperforms other methods, achieving the highest mean AUC score of $63.2\%$. Notably, AFiRe surpasses the previous best-performing methods, PCRL and DiRA, which obtain mean AUC scores of $62.9\%$ and $62.6\%$, respectively. 
With a $10\%$ labeling ratio, AFiRe continues to lead, attaining a mean AUC of $76.1\%$, outperforming PCRL ($75.8\%$) and DiRA ($74.9\%$). AFiRe shows significant improvements in $10$ categories, such as Fibrosis (Fibr.), Nodule (Nodu.), and Pleural Thickening (P.T.), with AUC scores of $71.9\%$, $70.5\%$, and $75.6\%$, respectively. 
\textit{These results underline the robustness of our approach in scenarios with limited labeled data to achieve enhanced performance in diagnosing small-sized lesions compared to others like Cardiomegaly or Effusion, which typically involve larger areas. }
In the fully labeled scenario ($100\%$) in Table \ref{tab:Transfer_expand}, AFiRe achieves the highest mean AUC score, marginally outperforming other radiography-specific SSL methods. 
These results confirm the efficacy of AFiRe in achieving state-of-the-art performance.

\textbf{\textit{Results with limited labeling on CXP Datasets.}} Right of Table \ref{tab:Transfer_expand} showcases the superior performance of the proposed AFiRe model, particularly at lower labeling ratios, highlighting its efficacy in data-scarce scenarios. Specifically, for the labeling ratio of $1\%$, our proposed model, AFiRe, achieves the highest mean AUC score of $78.8\%$, outperforming the closest competitor, DiRA, which scored $78.4\%$ ($0.4\% \downarrow$). When the labeling ratio increases to $10\%$, AFiRe maintains its leading position with a mean AUC score of $86.3\%$. This improvement is consistent across all disease labels, with AFiRe achieving top scores in Atelectasis ($81.6\%$) and matching the highest scores in Pleural Effusion ($90.6\%$). \textit{This performance underscores AFiRe's effectiveness in diagnosing small-scale and various-density lesions under limited labeled data.} At the full labeling ratio ($100\%$) in Table \ref{tab:Transfer_expand}, AFiRe continues to outperform other models. In particular, AFiRe outperforms the closely related methods (DiRA, Adamv1 and Adamv2) with an improved mean AUC of $+2.0\%$, $+1.9\%$, and $+1.0\%$, respectively. This result demonstrates the robustness of AFiRe. 

\textbf{\textit{Five-fold Cross-validation Results:}} We present the average AUC score across five iterations for (i) each of the 14 diseases within NIH datasets (see Table \ref{tab:NIH_disease}); (ii) each of the 5 diseases within CXP dataset (see Table \ref{tab:CXP_disease}). The mean AUC score for all diseases across each iteration is summarized in Table \ref{tab:5_mean}. Additionally, we analyze the statistical significance of our results compared to other methods, as illustrated in Fig. \ref{fig:p_value} (g) and (h). 
From these results, our AFiRe consistently achieves superior performance compared with the fully-supervised ImageNet model ($+4.6\%$ and $+3.7\%$ on NIH and CXP datasets respectively), as well as significant performance boosts ($p<0.05$) compared with all SSL counterparts. 
Specifically, for the NIH dataset, AFiRe shows an increased performance of 4.2\% and 2.8\% compared to the closest self-supervised baselines, DINO and MAE, respectively. Furthermore, AFiRe demonstrates an increased performance of 2.4\%, 1.5\%, and 1.2\% compared to the state-of-the-art radiography-specific self-supervised models C2L, DiRA, and PCLR, respectively. For the CXP datasets, AFiRe improves performance by 2.9\% and 2.5\% compared to DINO and MAE. Besides, AFiRe shows a highly significant improvement over C2L ($p<0.01$), DiRA ($p<0.01$), Adamav1 ($p<0.001$), and Adamv2 ($p<0.01$).
These results suggest that AFiRe is an effective self-supervised framework that promotes the learning of robust representations. It incorporates comprehensive anatomical discrimination and retains fine-grained information from images, enabling models to generalize across various medical tasks.

\section{H.More Anomaly Detection Analysis}
\textit{\textbf{Comparison setting:}}
The comparison methods for anomaly detection encompass two categories:  (i) the latest unsupervised anomaly detection methods designed for natural images, including MemAE, CutPast, PANDA, and M-KD; and (ii) the state-of-the-art unsupervised anomaly detection methods designed for medical images, including SALAD, f-AnoGAN, SQUID, and AnatPaste. Specifically, SALAD leverages GANs to capture the manifold of normal data, reconstructing abnormal images by introducing pixel corruption and shuffling to the original normal images.
f-AnoGAN detects anomalies by capturing intricate variations within normal images. 
SQUID uses space-aware memory queues for inpainting and detecting anomalies in radiographic images. AnatPaste illustrates the feasibility of training models with synthetic data for anomaly detection. For a fair comparison, we evaluate the results of \textit{child}CXR using ACC, AUC, and F1 scores reported in the respective papers. For the CXP dataset, we re-implemented AnatPaste due to its use of a different data partition. For the NIH dataset, we re-implemented all the models.
We report the original average AUC score and the standard deviation (SD) when available; otherwise, we repeat the experiments three times. The statistical significance of these results is shown in Fig. \ref{fig:p_value} (a-f).

\textit{\textbf{Results:}} For these three datasets, we provide statistical analysis of both ACC and AUC using an independent two-sample t-test. Specifically, on the \textit{Child}CXRs, AFiRe achieves the highest average scores with an ACC of $83.9\%$ and an F1 score of $87.6\%$, significantly surpassing the anatomy-based anomaly detection model, SQUID, with the $p<0.05$. Although AnatPaste slightly outperforms AFiRe in AUC by $0.6\%$, AFiRe shows a marked improvement in ACC ($p<0.05$)  and F1 scores, demonstrating its overall robustness.
In the CXP dataset, AFiRe outperforms the previous best model, recording the highest average scores in both ACC ($72.4\%$) and AUC ($79.9\%$). 
In the NIH dataset, AFiRe surpasses other models with an ACC of $66.8\%$, an AUC of $68.2\%, $ and an F1 score of $60.4\%$. While SQUID and AnatPaste tie in ACC and AUC, their F1 scores are slightly lower than AFiRe.
Overall, AFiRe achieves a significant improvement ($p<0.05$) across all tasks compared to all related methods, indicating that it performs better in terms of both overall accuracy and the balance between precision and recall.

\end{document}